%% file: main.tex
\documentclass[letterpaper]{article} 
\usepackage{aaai2027}  
\usepackage{newtxmath}
\usepackage[hyphens]{url}  
\usepackage{graphicx} 
\usepackage{natbib}  
\usepackage{caption} 
\usepackage{booktabs}
\usepackage{amsmath}
\usepackage{multirow}
\usepackage{array}
\usepackage{siunitx}
\usepackage[capitalize,noabbrev]{cleveref}
\newcommand{\lbl}{\textsc{Label}}
\crefname{appsec}{Appendix}{Appendices}
\Crefname{appsec}{Appendix}{Appendices}
\crefname{appsubsec}{Appendix}{Appendices}
\Crefname{appsubsec}{Appendix}{Appendices}
\nocopyright 
\title{\lbl{}: Literature Asymmetry between Efficacy and Limits\\in Drug Knowledge and Language Models}
\author{S~M~Rayeed, Charles~V.~Stewart, Yao~Ma}
\affiliations{Rensselaer Polytechnic Institute, Troy, NY  \qquad \{rayees, stewart, may13\}@rpi.edu}

\begin{document}

\maketitle

\input{Sections/0.abstract}

\input{Sections/1.intro}

\input{Sections/2.related_work}

\input{Sections/3.data_setting}

\input{Sections/4.methodology}

\input{Sections/5.results}

\input{Sections/6.dislimcon}

\bibliography{main}

\clearpage
\appendix

\crefalias{section}{appsec}
\crefalias{subsection}{appsubsec}

\input{Appendix/0_contents}

\input{Appendix/A_overview}

\input{Appendix/B_data}

\input{Appendix/C_corpora}

\input{Appendix/D_matching}

\input{Appendix/E_stats}

\input{Appendix/F_withindrug}

\input{Appendix/G_label}

\input{Appendix/H_probe}

\input{Appendix/I_page}

\input{Appendix/J_relcheck}

\input{Appendix/K_routing}

\input{Appendix/L_hetionet}

\input{Appendix/M_cases}

\end{document}

%% file: Sections/0.abstract.tex
\begin{abstract}

Every medication comes with two types of medical facts: \emph{indications}, which describe what conditions the drug treats (benefits) and \emph{contraindications}, which describe who must avoid it to prevent serious harm (safety limits). Although contraindications are critical for patient safety and FDA regulations require listing them as rigorously as indications, we find that contraindications are far less documented than indications in general prose. In this paper, we measure how often these texts contain both facts across \num{574} drugs from PrimeKG by comparing a drug's indications against its own contraindications. Across Wikipedia articles and PubMed abstracts, we find that reference prose names a drug's benefits about twice as often as its safety limits (within-drug odds ratio \num{2.19}). We call this documentation gap \lbl{} (Literature Asymmetry between Efficacy and Limits). Language models mirror this gap. As we scale Qwen2.5 from \num{0.5}B to \num{14}B, accuracy on indications rises by \num{18.3} points, whereas accuracy on contraindications remains flat near random guessing. However, when we provide models with a retrieved passage that states the answer, they use safety facts just as effectively as benefits. Furthermore, routing safety questions to official FDA labels boosts accuracy by \num{15.8} points whenever the label states the fact. Overall, our findings show that models fail on medication safety not because they cannot reason about risk, but because free-text corpora lack safety facts. Clinical AI systems should therefore route medication-safety questions to official regulatory labels and structured databases rather than relying on general public records. Code and data are available at \url{https://github.com/SMRayeed/LABEL}.

\end{abstract}

%% file: Sections/1.intro.tex
\section{Introduction}

If we look up a medication online, its Wikipedia page usually tells us what it treats, but it is much less likely to mention who must never take it. In medicine, the condition a drug treats is called an \emph{indication}. For example, metformin treats type 2 diabetes. Conversely, a condition that makes a drug dangerous to take is called a \emph{contraindication}. For example, ciprofloxacin, a common antibiotic, must not be given to people with myasthenia gravis, a disease that weakens the muscles, because the drug can worsen muscle weakness. Structured clinical databases such as DrugCentral record both kinds of facts \citep{ursu2017drugcentral}. However, free text records like Wikipedia or PubMed abstracts do not document them equally. This gap matters because large language models (LLMs) learn from these text corpora, and patients and clinicians increasingly ask the LLMs about medications. If the corpora record contraindications less often than indications, LLMs may be less reliable on questions where a wrong answer can cause patients the greatest harm. Recognizing this gap also changes how we diagnose and fix the issue. When a model misses a contraindication, the model itself is not necessarily bad at reasoning about safety; rather, the safety fact may simply be missing from the prose on which it was trained. And in that case, writing better prompts or building better retrievers will not help. Instead, clinical systems must route safety questions to structured databases and regulatory documents that record contraindications.

\begin{figure*}[t]
\centering
\includegraphics[width=\textwidth]{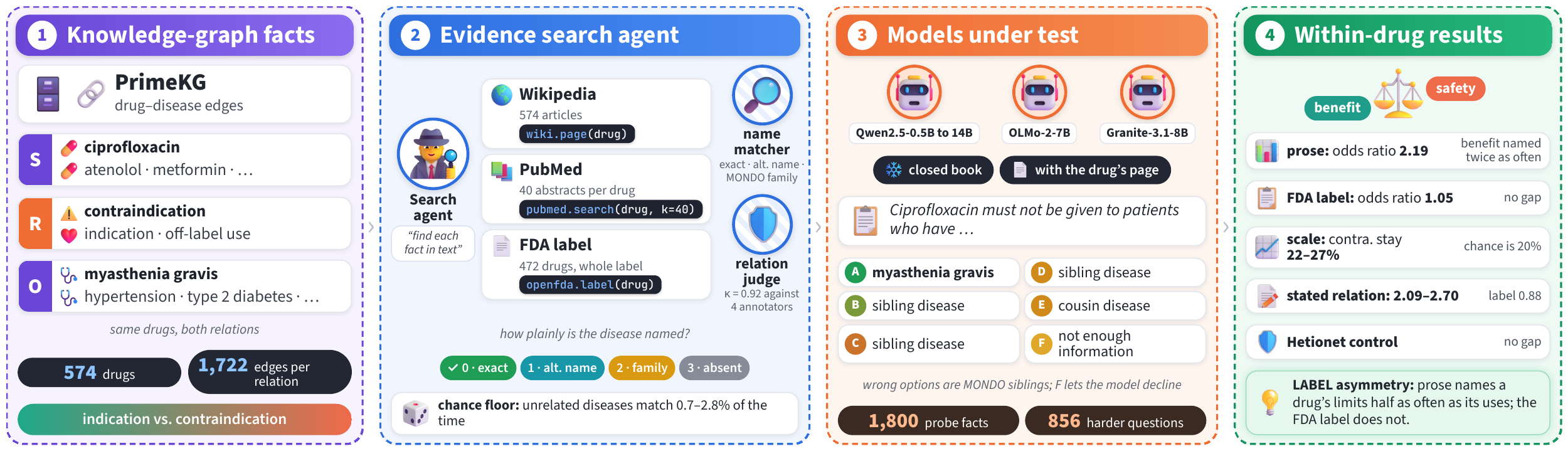}
\caption{Overview of our study pipeline. (\num{1}) We sample paired benefit facts (\emph{indications}) and safety facts (\emph{contraindications}) for the exact same drugs from the PrimeKG knowledge graph. (\num{2}) We retrieve text from Wikipedia, PubMed abstracts, and official FDA drug labels, and check both whether the text mentions the target disease and whether the sentence states the correct medical relationship. (\num{3}) We test language models on the same facts both closed book (from memory alone) and open book (with a retrieved passage). (\num{4}) Comparing benefit and safety facts within each drug reveals the \lbl{} asymmetry.}
\label{fig:pipeline}
\end{figure*}

Prior works in two research directions explored pieces of this problem. First, pharmacy and medical informatics studies audit reference websites like Wikipedia by checking a small set of drugs against clinical textbooks \citep{clauson2008scope,kraenbring2014accuracy}. Second, AI benchmarks built from biomedical knowledge graphs (structured networks that store facts as subject-relation-object links) test how well LLMs answer clinical questions \citep{liu2026mhgraphbench}, but they do not check whether the underlying facts actually appear in text corpora. Moreover, neither line of work controls for drug popularity: widely prescribed drugs have longer articles and thousands of research papers, so comparing facts across different drugs mixes a drug's overall popularity with the type of fact being measured. In this work, we bridge these two directions by evaluating text coverage and model knowledge on the same drug-disease facts (\cref{fig:pipeline}). Our main contributions are as follows:

\begin{itemize}
    \item \textbf{Within-drug text audit.} We design a paired, within-drug evaluation across \num{574} drugs from PrimeKG, comparing how often text documents each drug's indications versus its own contraindications so that article length, drug fame, and research volume cancel out.
    \item \textbf{Cross-corpus audit.} We evaluate three text sources: English Wikipedia articles, PubMed research abstracts (using a balanced search budget for benefits and harms), and official FDA drug labels. We show that both Wikipedia and PubMed state benefits more than twice as often as safety limits, whereas FDA labels document both equally. We also verify with a language-model judge (Qwen2.5-7B) and four human annotators that matched sentences genuinely mention the medical relationship rather than just mentioning the disease.
    \item \textbf{Scaling and distractor difficulty.} We test \num{1800} drug-disease facts across five Qwen2.5 models from \num{0.5}B to \num{14}B parameters. Scaling improves indication accuracy by \num{18.3} points but contraindication accuracy by only \num{5.0}. We repeat the test with harder, closely related wrong options in four models from three families (Qwen2.5, OLMo-2 and Granite-3.1), and the gap holds in every model.
    \item \textbf{Separating retrieval from reading.} We test models with retrieved passages and separate performance by whether a passage explicitly contains the target condition. We show that when a passage states a contraindication, LLMs use it just as well as they use a stated indication.
    \item \textbf{Negative control and clinical routing.} We validate our pipeline on Hetionet, a negative control whose two drug-disease relations both describe benefit, and find no gap. Finally, we demonstrate a relation-aware routing strategy that redirects safety questions to the FDA labels, boosting accuracy by \num{15.8} points when the label states the fact.
\end{itemize}

%% file: Sections/2.related_work.tex
\section{Related Work}


\paragraph{Drug information in reference text.}
Wikipedia is one of the most widely read health resources online \citep{heilman2011wikipedia}, and pharmacy studies have audited its drug articles for more than a decade. \citet{clauson2008scope} compare Wikipedia with a professional drug compendium and find a narrower scope and more omissions. \citet{kraenbring2014accuracy} compare \num{100} drug articles with pharmacology textbooks and find high accuracy but incomplete coverage. Other audits compare Wikipedia with Micromedex, Lexicomp and FDA medication guides, and find it less complete \citep{lavsa2011reliability,reilly2017accuracy,candelario2017completeness,hunter2018comparison}. Two findings are close to ours. \citet{kupferberg2011accuracy} examine the Wikipedia articles of five statins, a class of cholesterol-lowering drugs, and find no incorrect statements but frequent omissions of contraindications and drug interactions. \citet{hwang2014drugsafety} find that more than a third of drug articles are not updated within a year of an FDA safety warning. These audits cover at most about a hundred drugs, grade articles against a reference text rather than a drug's full set of facts, and do not link missing facts to model errors.

\paragraph{Structured drug databases and knowledge graphs.}
Unlike general prose, regulatory and clinical databases are built specifically to track drug safety. DrugCentral curates approved indications and contraindications from regulatory records \citep{ursu2017drugcentral}, SIDER extracts adverse drug reactions from package inserts \citep{kuhn2016sider}, and openFDA provides programmatic access to the full text of official FDA drug labels \citep{kasshout2016openfda}. Biomedical knowledge graphs such as PrimeKG and Hetionet integrate these curated databases with standardized disease ontologies \citep{chandak2023primekg,himmelstein2017hetionet}. We use these structured graphs as ground truth to test, fact by fact, how much of this clinical knowledge actually makes it to the free text.

\paragraph{Medical knowledge in LLMs.}
LLMs achieve high scores on standard medical licensing exams \citep{singhal2023clinical,singhal2025expert}. However, evaluations built from biomedical knowledge graphs reveal that model performance varies sharply depending on the type of clinical relation. For instance, MHGraphBench \citep{liu2026mhgraphbench} tests models on questions derived from PrimeKG and finds that predicting specific relations between entities is much harder than identifying entity types. While existing benchmarks document what LLMs get wrong, they do not trace those failures back to what is missing in reference text.

\paragraph{Corpus frequency and model recall.}
A separate line of research shows that LLMs remember facts better when facts appear frequently in pretraining text. \citet{kandpal2023longtail} demonstrate that question-answering accuracy scales with number of relevant documents in the training corpus, and \citet{mallen2023trust} show that LLMs struggle with rare entities unless assisted by retrieval. Similarly, LMEnt \citep{gottesman2026lment} indexes entity mentions in pretraining data to study how models acquire factual knowledge. However, these studies measure popularity at the level of an entire entity (such as a drug name). Because a drug's benefits and safety limits share the same drug name, entity-level popularity cannot explain why a model knows what a popular drug treats but not who must avoid it. We resolve this by measuring coverage at the level of individual facts and comparing benefit versus safety relations within the same drug.

\paragraph{RAG in medicine.}
Retrieval-augmented generation (RAG) is a standard way to help models answer questions when parametric memory fails \citep{lewis2020rag}. In biomedicine, the MIRAGE benchmark \citep{xiong2024medrag} shows that retrieving passages from PubMed, textbooks, and Wikipedia improves overall accuracy. Yet existing medical RAG benchmarks report aggregate accuracy across entire question sets without separating indication questions from contraindication questions and without checking whether the retrieved text actually states the answer. Recent works on context sufficiency \citep{joren2025sufficient, rayeed2026clade} show that RAG helps only when the retrieved passage contains the needed fact. We evaluate this condition for drug safety and show that retrieval fails far more often for contraindications than for indications. Our findings complement recent systems like \citet{bang2025contraindications}, which answer contraindication queries by querying structured drug databases.

\paragraph{How data limits what models learn.}
Our investigation builds on a broader observation from our prior work in scientific AI: gaps in underlying data limit what downstream models can learn. In Bio-VQA, model accuracy drops from \num{0.72} at phylum to \num{0.12} at species \citep{rayeed2027biovqa}. We observe similar granularity drops in beetle classification \citep{rayeed2025beetleverse,rayeed2026beetles}, where specimen imaging protocols constrain what computer vision models can extract \citep{east2025imaging}. Likewise, knowledge-based VQA benchmarks often include questions that their own reference sources fail to support \citep{ma2026pitfalls}, and targeted evaluation sets regularly expose blind spots hidden by broad benchmarks \citep{stevens2025mindgap}. Here, we ask the same question for medication safety: what does reference text actually record before any LLM reads it?

%% file: Sections/3.data_setting.tex
\section{Data and Study Setup}

\input{tables/tab_data}

\paragraph{Three types of drug-disease facts.}
We study three clinical relationships between a drug and a disease. An \emph{indication} is a disease that the drug is officially approved to treat. A \emph{contraindication} is a condition in which the drug must not be given because the risk of patient harm outweighs any possible benefit. An \emph{off-label use} is a condition that doctors treat with the drug in clinical practice even though the use is not part of its formal regulatory approval. Indications and off-label uses describe therapeutic benefits, whereas contraindications define safety limits that protect patients from harm.

\paragraph{Knowledge graphs and disease ontologies.}
We draw our primary drug-disease facts from PrimeKG \citep{chandak2023primekg}, a large biomedical knowledge graph that integrates \num{20} clinical resources. In a knowledge graph, each fact is stored as a link, called an \emph{edge}. An edge connects a drug node to a disease node under a specific relation. PrimeKG takes its drug-disease edges from DrugCentral \citep{ursu2017drugcentral}, comprising \num{9388} indication edges, \num{30675} contraindication edges, and \num{2568} off-label use edges. Every disease in PrimeKG is linked to the MONDO Disease Ontology \citep{vasilevsky2022mondo}, a tree-like hierarchy that organizes conditions from broad categories (such as \emph{diabetes mellitus}) down to specific subtypes (\emph{type 1 diabetes mellitus}). Removing overly broad top-level categories with more than \num{400} descendants leaves \num{17801} parent-child links across \num{11788} diseases. As a negative control, we evaluate Hetionet \citep{himmelstein2017hetionet}, an independent biomedical knowledge graph built from different underlying sources. Hetionet contains \num{755} \emph{treats} edges (curative therapy) and \num{390} \emph{palliates} edges (where a drug relieves symptoms without curing the underlying disease). Because both relations in Hetionet describe therapeutic benefits, it serves as a control to verify that our text-matching pipeline does not manufacture artificial gaps between relations.

\paragraph{Text corpora.}
We test how often drug-disease facts appear across three distinct text sources:
\begin{enumerate}
\item \textbf{English Wikipedia:} We retrieve the full English Wikipedia article for each drug, representing general-purpose reference prose written for a broad audience.
\item \textbf{PubMed abstracts:} We retrieve \num{40} relevance-ranked biomedical research abstracts per drug via the NCBI E-utilities API \citep{sayers2009eutils}. To make sure our search does not favor treatment papers over safety papers, we split the query budget evenly: we retrieve \num{20} abstracts using a therapeutic-use query and \num{20} abstracts using an adverse-effect and contraindication query.
\item \textbf{Official FDA drug labels:} We retrieve official manufacturer prescribing labels from openFDA \citep{kasshout2016openfda}. We search for drugs by generic name, substance name, and brand name, pooling up to five labels per drug so that medications sold by multiple manufacturers are thoroughly covered. From each label, we extract three text views: the \emph{Indications and Usage} section alone, the \emph{Contraindications} section alone, and the \emph{whole label} (which includes all sections, such as the Boxed Warning, Warnings and Precautions, Adverse Reactions, and Use in Specific Populations). We find official FDA labels for \num{472} of our \num{574} sampled drugs (\num{82.2}\%).
\end{enumerate}

\paragraph{Evaluation samples.}
To compare indications and contraindications within the same drug, we identify all \num{595} drugs in PrimeKG that have at least three indication edges and at least three contraindication edges, and we randomly sample exactly three of each per drug. Among these, \num{574} drugs have an English Wikipedia article of sufficient length, yielding a balanced text-audit sample of \num{1722} indication facts and \num{1722} contraindication facts. The \num{472} drugs with an FDA label provide \num{1416} matched facts per relation. For our closed-book model scaling experiments, we draw a separate sample of \num{600} facts for each of the three relations (\num{1800} facts total). \Cref{tab:data} summarizes all datasets and sample sizes.

%% file: tables/tab_data.tex
\begin{table}[t]
\centering
\small
\setlength{\tabcolsep}{4pt}
\begin{tabular}{llr}
\toprule
Source & Item & Count \\
\midrule
PrimeKG & indication edges & \num{9388} \\
 & contraindication edges & \num{30675} \\
 & off-label use edges & \num{2568} \\
Hetionet & treats / palliates edges & \num{755} / \num{390} \\
MONDO & parent links / diseases & \num{17801} / \num{11788} \\
\midrule
Text sample & drugs & \num{574} \\
 & edges per relation & \num{1722} \\
FDA label sample & drugs with a label & \num{472} \\
 & edges per relation & \num{1416} \\
Probe sample & edges per relation & \num{600} \\
 & model sizes (Qwen2.5) & \num{5} \\
Page test & ind.\ / contra.\ questions & \num{450} / \num{406} \\
 & models (families) & \num{4} (\num{3}) \\
\bottomrule
\end{tabular}
\caption{Data in this study. An \emph{edge} is one drug-disease fact in a knowledge graph. The lower block gives the samples we draw for each test.}
\label{tab:data}
\end{table}

%% file: Sections/4.methodology.tex
\section{Methodology}

We run four connected experiments on the same drug facts. We (1) measure how often each corpus names each fact, (2) compare benefit coverage against safety coverage within each drug, (3) test what LLMs know from memory across model sizes and question difficulty, and (4) test how much a retrieved passage from prose or the FDA label helps.

\subsection{Measuring Whether Text States a Fact}

\input{tables/tab_levels}

\paragraph{Granularity of disease mentions.}
For each drug-disease edge, we check whether the drug's text mentions the disease at three levels (\cref{tab:levels}). Level \num{0} is the disease's exact MONDO label (\emph{type 1 diabetes mellitus}). Level \num{1} adds synonyms, lay terms and abbreviations taken from the titles of Wikipedia pages that redirect to the disease's article (\emph{juvenile diabetes}). Level \num{2} adds the names of the disease's MONDO ancestors up to two levels above it (\emph{diabetes}). The matcher ignores case, respects word boundaries so that \emph{asthma} does not match \emph{asthmatic}, and skips names shorter than four characters. We apply the same matcher to Wikipedia, PubMed and FDA labels (\cref{app:matching}).

\paragraph{What counts as a named fact.}
We count a fact as \emph{named} only at Level \num{0} or Level \num{1}. A family mention (Level \num{2}) does not count, because saying that a drug is risky in \emph{liver disease} does not identify a specific condition such as \emph{cirrhosis}.

\paragraph{Verifying medical relationships in text.}
A disease name in a drug's article does not guarantee that the sentence states the clinical relationship. We extract the first sentences that name the disease and ask an instruction-tuned judge (Qwen2.5-7B) to classify each into one of four categories: (A) the drug treats or manages the disease, (B) the drug must not be used in patients with the disease, (C) the drug may cause or worsen the disease or requires caution, or (D) the sentence names both without a clinical link. A sentence supports an indication when the judge selects (A), and a contraindication when it selects (B) or (C). Four human annotators independently labeled \num{160} matched sentences with the graph relation hidden, to validate the judge (\cref{app:relcheck}).

\paragraph{Control for random matches.}
A disease name may appear in a drug's article by coincidence. To measure this \emph{random floor}, we pair each drug with three unrelated diseases from the same relation and run the same matcher. On Wikipedia, unrelated diseases match in only \num{1.22}\% of indication pairs, \num{0.72}\% of off-label pairs and \num{0.78}\% of contraindication pairs. Full FDA labels are much longer, so their random floor is higher (\num{2.10}\% and \num{2.82}\%), but still an order of magnitude below real match rates.

\subsection{Comparing Each Drug Against Itself}

\paragraph{Within-drug odds ratio.}
Commonly prescribed drugs have long Wikipedia articles, dozens of PubMed papers and extensive FDA labels. Comparing indications from famous drugs against contraindications from obscure drugs would confound relation type with drug popularity. We remove this confounder by comparing indications and contraindications \emph{within each drug}. For each drug $d$, we form a 2$\times$2 table of its named indications ($I^{+}_{d}$), unnamed indications ($I^{-}_{d}$), named contraindications ($C^{+}_{d}$) and unnamed contraindications ($C^{-}_{d}$), where $N_d$ is its number of sampled facts. We pool these tables over the set of drugs $\mathcal{D}$ with Mantel-Haenszel odds ratio, $\mathrm{OR}_{\mathrm{MH}}$ (\citealt{mantel1959}):
\begin{equation}
\mathrm{OR}_{\mathrm{MH}}
  = \frac{\sum_{d \in \mathcal{D}} w_d\, \mathrm{OR}_d}{\sum_{d \in \mathcal{D}} w_d},
\quad
\mathrm{OR}_d = \frac{I^{+}_{d}\, C^{-}_{d}}{I^{-}_{d}\, C^{+}_{d}},
\quad
w_d = \frac{I^{-}_{d}\, C^{+}_{d}}{N_d}.
\label{eq:mh}
\end{equation}
The \emph{odds} of a fact being named are the probability that the text names it divided by the probability that it does not. $\mathrm{OR}_{\mathrm{MH}}$=\num{1} means that a drug's benefits and safety limits are named at equal rates. $\mathrm{OR}_{\mathrm{MH}}$=\num{2} means that, for the same drug, the odds of naming a benefit are twice the odds of naming a safety limit. The Cochran-Mantel-Haenszel test gives the $p$-value.

\paragraph{Control for the number of facts per drug.}
In PrimeKG, drugs have about \num{3}$\times$ as many contraindication edges as indication edges. A drug with dozens of contraindications may have some omitted simply because no summary lists every risk. We control for this in two ways. First, a logistic regression predicts whether a fact is named from its relation type, holding fixed the logarithm of the drug's edge count in that relation. Second, we group drugs by their edge count in the relation (\num{3} to \num{5}, \num{6} to \num{12}, and \num{13} or more) and measure the gap within each group.

\subsection{Testing What LLMs Know}

\paragraph{Closed-book knowledge probe.}
To test what LLMs know without retrieved text (\emph{closed book}), we turn the \num{1800} probe facts into five-option questions. Each question pairs the true disease with four random diseases from the same relation that have no link to the drug. We append each option to a relation stem, such as \emph{``X is a drug used to treat the disease called''}, and pick the option with the highest mean token log-probability. Random guessing gives \num{20}\% accuracy. We evaluate five instruction-tuned Qwen2.5 models: 0.5B, 1.5B, 3B, 7B and 14B \citep{qwen2025qwen25} (\cref{app:probe}). Because all five models answer the same questions, we compare their gains from scale fact by fact, and test the difference with a paired bootstrap over drugs.

\subsection{Harder Distractors, Open-Book Retrieval, and Routing}

\paragraph{Harder options with an abstention choice.}
Random distractors can sometimes be ruled out by basic medical intuition. We therefore build a harder set of \num{450} indication and \num{406} contraindication questions whose four wrong options are MONDO siblings of the true disease first, then cousins. A sixth option, \emph{``Not enough information to tell''}, lets the model abstain instead of guessing. We stratify the questions (up to \num{150} per group and relation) by whether the drug's prose passage names the exact disease, names only its family, or is silent. We evaluate Qwen2.5-0.5B, Qwen2.5-7B, OLMo-2-7B, and Granite-3.1-8B models.

\paragraph{Open-book evaluation with retrieved prose.}
We ask the \num{856} harder questions twice: once closed book, and once with a retrieved passage in the prompt. The passage is a cleaned excerpt of at most \num{1800} characters from the drug's Wikipedia article and PubMed abstracts, built around any mention of the disease or its family (\cref{app:page}).

\paragraph{Routing questions to FDA labels.}
Finally, we test whether routing safety questions to the FDA label helps. For the \num{726} questions whose drug has a label, we cut a passage of the same length from the whole label. The \emph{routed} setup uses the prose passage for indication questions and the label passage for contraindication questions (\cref{app:routing}).

%% file: tables/tab_levels.tex
\begin{table}[t]
\centering
\small
\setlength{\tabcolsep}{4pt}
\begin{tabular}{lll}
\toprule
Level & Text contains & Example \\
\midrule
\num{0} & exact name & ``type 1 diabetes mellitus'' \\
\num{1} & an alternative name & ``juvenile diabetes'' \\
\num{2} & only the family & ``diabetes'' \\
\num{3} & nothing & - \\
\bottomrule
\end{tabular}
\caption{Levels at which a text can name a disease. Levels \num{0} and \num{1} count as \emph{named}.}
\label{tab:levels}
\end{table}

%% file: Sections/5.results.tex
\section{Results and Analysis}

\subsection{Text Favors Benefits 2$\times$ over Safety Facts}
\label{sec:results-coverage}

\begin{figure}[t]
\centering
\includegraphics[width=\columnwidth]{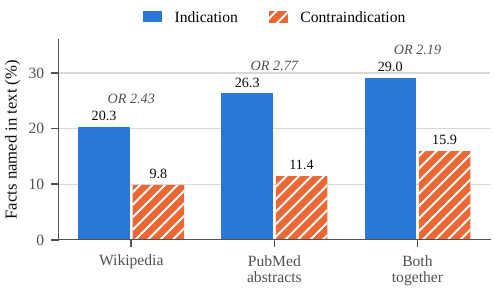}
\caption{Percentage of facts named in each prose corpus for the same \num{574} drugs (\num{1722} facts per relation). Numbers above each pair of bars give the within-drug odds ratio ($\mathrm{OR}_{\mathrm{MH}}$); values above \num{1} mean the corpus names benefits more often than safety limits.}
\label{fig:coverage}
\end{figure}

\input{tables/tab_coverage}

\paragraph{The \lbl{} gap.}
Wikipedia names \num{20.33}\% of a drug's indications but only \num{9.81}\% of its contraindications (\cref{tab:coverage,fig:coverage}). Comparing each drug against itself, the odds that Wikipedia names a benefit are \num{2.43} times the odds that it names a safety limit. PubMed abstracts show a wider gap, \num{26.31}\% against \num{11.44}\%, with an odds ratio of \num{2.77}. Even with Wikipedia and PubMed combined, the odds ratio stays at \num{2.19}. A reader or retriever that checks both sources finds nearly three in ten indications (\num{29.04}\%) but fewer than one in six contraindications (\num{15.91}\%).

\paragraph{Dedicated searches do not close the contraindication gap.}
Our PubMed search spends half of its budget on adverse effects and contraindications. Compared with a single general query of \num{40} abstracts, this balanced search raises contraindication coverage by only \num{1.22} points (\num{10.22}\% to \num{11.44}\%) and indication coverage by \num{2.97} points (\num{23.34}\% to \num{26.31}\%). The odds ratio is \num{2.77} under both searches. Toxicity searches retrieve safety papers, but those papers rarely name the specific conditions that rule a drug out.

\begin{figure}[t]
\centering
\includegraphics[width=\columnwidth]{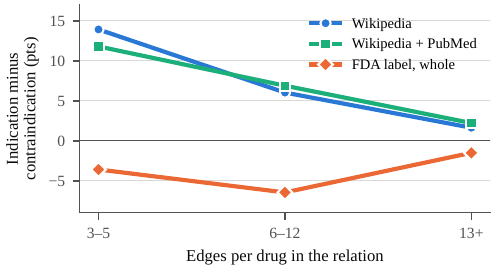}
\caption{Coverage gap (indication minus contraindication naming rate, in points) grouped by the drug's number of facts in that relation. Positive bars mean the text favors indications. Prose favors indications in every group, whereas the FDA label does not.}
\label{fig:degree}
\end{figure}

\paragraph{The safety gap is widest for drugs with few facts.}
\Cref{fig:degree} splits the gap by how many facts a drug has in PrimeKG. In Wikipedia, the gap is \num{+13.94} points for drugs with \num{3} to \num{5} facts, \num{+6.04} points for \num{6} to \num{12} facts, and \num{+1.64} points for \num{13} or more. Wikipedia and PubMed combined show the same pattern (\num{+11.81}, \num{+6.90} and \num{+2.23} points). With the edge count held fixed in the logistic model, contraindications remain less likely to be named ($p$=\num{2.4e-4} in Wikipedia, \num{1.4e-5} in PubMed and \num{9.7e-4} combined).

\paragraph{Benefits are often written at the family level, safety limits are not.}
Relaxing the matcher shows where the missing mentions go. On Wikipedia, alternative names add only \num{1.39} points on average. Family names up to two levels add \num{14.06} points for indications but only \num{1.72} points for contraindications, after subtracting the random floor. Prose often says that a drug treats a broad group of diseases, but it rarely names a contraindication even at the family level.

\input{tables/tab_relcheck}

\paragraph{Matched sentences state clinical relationships.}
\Cref{tab:relcheck} counts a fact only when its sentence states the relation, as read by the Qwen2.5-7B judge. The four annotators agree closely with one another (Fleiss' $\kappa$=\num{0.89}), and the judge matches their majority vote on \num{95.5}\% of sentences (Cohen's $\kappa$=\num{0.92}), where $\kappa$=\num{1} is perfect agreement and $\kappa$=\num{0} is chance. Under this stricter count, the prose gap remains (odds ratios of \num{2.70} in Wikipedia and \num{2.09} in PubMed), and the whole FDA label remains balanced (\num{0.88}). The annotators also show how safety facts are phrased: \num{82}\% to \num{83}\% of matched contraindication sentences are warnings or precautions, and only \num{10}\% to \num{11}\% are formal prohibitions. This is why a keyword filter agrees poorly with the annotators ($\kappa$=\num{0.43}).

\subsection{FDA Labels Show No Coverage Gap}

\begin{figure}[t]
\centering
\includegraphics[width=\columnwidth]{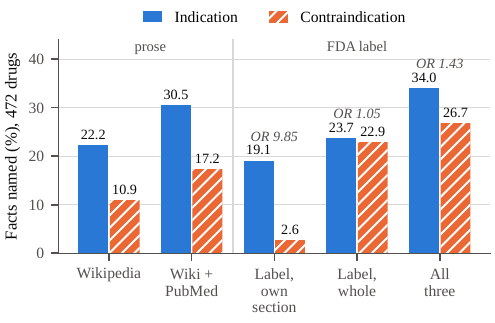}
\caption{Percentage of facts named for the \num{472} drugs with an FDA label (\num{1416} facts per relation). \emph{Own section} checks only \emph{Indications and Usage} for indications and only \emph{Contraindications} for contraindications. Numbers above the bars give $\mathrm{OR}_{\mathrm{MH}}$. The whole label shows no gap ($\mathrm{OR}_{\mathrm{MH}}$=\num{1.05}).}
\label{fig:label}
\end{figure}

\input{tables/tab_label}

\paragraph{The FDA label names both relations.}
If contraindications were missing from prose because they are obscure edge cases, they would be missing from regulatory documents too. They are not. The whole FDA label names \num{23.73}\% of indications and \num{22.88}\% of contraindications on the same facts (\cref{fig:label,tab:label}), an odds ratio of \num{1.05} ($p$=\num{0.62}). With the edge count held fixed, the label slightly favors contraindications (coefficient \num{+0.30}, $p$=\num{0.005}).

\paragraph{Safety facts sit across the whole label.}
The \emph{Contraindications} section alone names only \num{2.61}\% of PrimeKG's contraindications, and it is blank for \num{9.5}\% of contraindication edges. Most safety facts sit in the \emph{Boxed Warning}, \emph{Warnings and Precautions} and \emph{Use in Specific Populations} sections. To check that the matcher did not miss reworded text, four annotators read \num{50} contraindications that the section failed to match. In \num{45} of the \num{49} cases with a majority vote, the condition is absent from the section, and only \num{3} state it in other words. A clinical retriever must therefore index the whole label, not just its \emph{Contraindications} section.

\paragraph{The label rescues safety facts that prose misses.}
Adding the whole label to Wikipedia and PubMed raises coverage to \num{34.04}\% for indications and \num{26.69}\% for contraindications, and the odds ratio falls to \num{1.43}. With the edge count held fixed, the gap disappears ($p$=\num{0.86}). Of the \num{1172} contraindications that both prose corpora miss, the label names \num{11.4}\%, more than twice its rate for missed indications (\num{5.1}\% of \num{984}). The \lbl{} asymmetry is therefore a property of general prose, not of the facts themselves.

\subsection{Scaling Helps Benefits, Not Safety Facts}

\begin{figure}[t]
\centering
\includegraphics[width=\columnwidth]{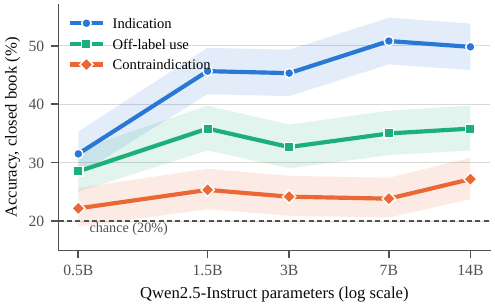}
\caption{Closed-book five-choice accuracy (\%) across five Qwen2.5 sizes (\num{0.5}B to \num{14}B) on \num{600} facts per relation. Shaded bands show \num{95}\% intervals, and the dashed line marks the \num{20}\% chance rate. Scale improves accuracy on indications but leaves contraindications near chance.}
\label{fig:scale}
\end{figure}

\paragraph{Closed-book accuracy across five model sizes.}
Across a \num{28}-fold increase in parameters, indication accuracy climbs from \num{31.50}\% at \num{0.5}B to \num{50.83}\% at \num{7}B and \num{49.83}\% at \num{14}B (\cref{fig:scale}; \cref{app:probe}). Contraindication accuracy stays between \num{22.17}\% and \num{27.17}\% at every size, barely above the \num{20}\% chance rate. Off-label uses fall in between, rising from \num{28.50}\% to \num{35.83}\%. At \num{7}B and \num{14}B, models answer about half of the indications correctly but do little better than chance on contraindications, a gap of \num{22.7} to \num{27.0} points. Paired fact by fact from Qwen2.5-0.5B to Qwen2.5-7B, indications gain \num{19.33} points and contraindications only \num{1.67} points ($p$=\num{4.8e-12}). From \num{0.5}B to \num{14}B, the gains are \num{18.33} points for indications, \num{7.33} for off-label uses and \num{5.00} for contraindications.

\paragraph{Accuracy tracks text coverage.}
The three relations rank in the same order in model accuracy as in prose coverage: indications first, off-label uses second and contraindications last. Split by whether Wikipedia names the fact, scaling from \num{0.5}B to \num{14}B improves accuracy by \num{17.5} to \num{26.7} points on named facts but only \num{4.1} to \num{16.6} points on unnamed facts. Wikipedia names only \num{40} of the \num{600} probe contraindications, against \num{101} of the \num{600} indications, so larger models have few written safety facts to learn from. Even among named facts, Qwen2.5-14B reaches \num{84.2}\% on indications but \num{40.0}\% on contraindications. One likely reason is that an article states a drug's uses up front, while its warnings sit in long lists.

\input{tables/tab_distractors}

\paragraph{The gap persists with harder options.}
With the MONDO siblings as wrong options and an abstention choice, every model shows a large closed-book gap favoring indications (\cref{tab:distractors}): \num{+7.9} points for Qwen2.5-0.5B, \num{+30.9} for Qwen2.5-7B, \num{+23.0} for OLMo-2-7B and \num{+45.9} for Granite-3.1-8B.

\begin{figure}[t]
\centering
\includegraphics[width=\columnwidth]{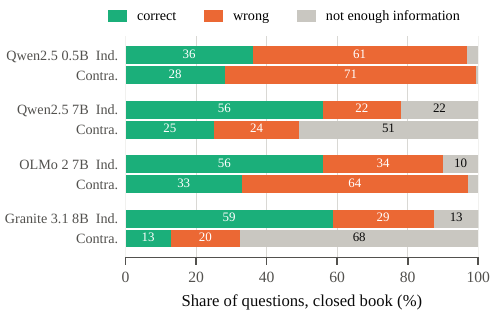}
\caption{Closed-book outcomes on the harder questions (\num{450} indication and \num{406} contraindication questions), split into correct answers, wrong guesses and abstentions (\emph{Declines}). On contraindications, Qwen2.5-0.5B and OLMo-2-7B mostly guess wrong, whereas Qwen2.5-7B and Granite-3.1-8B mostly abstain.}
\label{fig:failure}
\end{figure}

\paragraph{LLM failures: wrong guesses or abstention.}
The abstention option reveals two failure styles on contraindications (\cref{fig:failure}). Qwen2.5-0.5B and OLMo-2-7B rarely abstain and answer \num{71.4}\% and \num{64.0}\% of contraindication questions wrongly. Qwen2.5-7B and Granite-3.1-8B abstain on more than half of them (\num{50.7}\% and \num{67.5}\%), against only \num{22.0}\% and \num{12.7}\% of indication questions. Either way, precision on the contraindication questions a model does answer is \num{9} to \num{28} points lower than on indications (\cref{app:page}).

\subsection{RAG helps only when the fact is written}

\begin{figure}[t]
\centering
\includegraphics[width=\columnwidth]{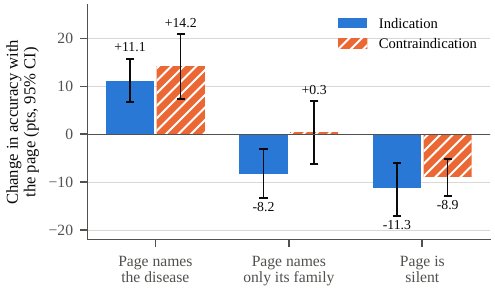}
\caption{Change in accuracy (points) when a retrieved prose passage is added to the prompt, averaged over Qwen2.5-7B, OLMo-2-7B and Granite-3.1-8B, grouped by whether the passage names the disease, names only its MONDO family, or is silent. When the passage names the disease, models gain as much on safety limits (\num{+14.2}) as on benefits (\num{+11.1}).}
\label{fig:retrieval}
\end{figure}

\paragraph{Models read written safety facts just as well as benefits.}
If LLMs fail on contraindication questions because they could not handle safety phrasing, a passage stating a safety fact would help less than one stating a benefit. It does not (\cref{fig:retrieval}). When the passage names the disease, accuracy rises by \num{14.2} points on contraindications and \num{11.1} points on indications across the three \num{7}B to \num{8}B models. Every model shows the same pattern (\cref{app:page}).

\paragraph{Silent passages versus family mentions.}
When a passage is silent, accuracy falls by \num{11.3} points on indications and \num{8.9} points on contraindications, because the models abstain more (\emph{Not enough information} rises by \num{26.0} and \num{32.9} points). This is the right behavior when the context lacks the answer. When a passage names only the disease family, the models abstain less than with a silent passage and often pick a sibling disease instead. Indication accuracy falls by \num{8.2} points, and contraindication accuracy does not change (\num{+0.3}).

\input{tables/tab_route}

\paragraph{Routing safety questions to FDA labels.}
\Cref{tab:route} replaces the prose passage with a label passage. On the \num{135} contraindication questions whose label names the condition, accuracy rises from \num{40.0}\% to \num{55.8}\%, a gain of \num{15.8} points, and abstentions fall by \num{12.1} points. When the label does not name it (\num{211} questions), the label passage lowers accuracy slightly (\num{-4.9} points). For indications, the label adds little when it names the disease (\num{+3.4}) and hurts when it does not (\num{-11.0}). Prose is the better source for benefit questions, and label is better for safety questions whose answer it states.

\subsection{Negative Control on Hetionet Shows No Gap}

Hetionet's \emph{treats} and \emph{palliates} relations both describe benefit, so a sound matcher should find no gap between them. On Wikipedia, it names \num{57.33}\% of \emph{treats} facts and \num{56.10}\% of \emph{palliates} facts, an odds ratio of \num{1.05} ($p$=\num{0.70}). Among the \num{51} drugs that have both relations, the within-drug odds ratio is \num{2.48}, but this is not significant once the edge count is held fixed ($p$=\num{0.16}), and its sign flips across edge-count groups (\cref{app:hetionet}). Under the same controls, the PrimeKG gap stays significant in every prose corpus. With the FDA label result, this control shows that the \lbl{} asymmetry comes from the prose, not from the measurement.

%% file: tables/tab_coverage.tex
\begin{table}[t]
\centering
\small
\setlength{\tabcolsep}{4pt}
\begin{tabular}{lcccc}
\toprule
Corpus & Ind. & Contra. & OR$_{\mathrm{MH}}$ & $p$ \\
\midrule
Wikipedia & \num{20.33} & \num{9.81} & \num{2.43} & \num{2e-18} \\
PubMed & \num{26.31} & \num{11.44} & \num{2.77} & \num{5e-29} \\
Both & \num{29.04} & \num{15.91} & \num{2.19} & \num{2e-20} \\
\bottomrule
\end{tabular}
\caption{Share of facts (\%) that text names, within-drug odds ratio of indication against contraindication and its Cochran-Mantel-Haenszel ($p$): 574 drugs, 1722 edges per relation.}
\label{tab:coverage}
\end{table}

%% file: tables/tab_relcheck.tex
\begin{table}[t]
\centering
\footnotesize
\setlength{\tabcolsep}{4pt}
\begin{tabular}{lccc}
\toprule
Corpus & Named & Stated & Share (ind./con.) \\
\midrule
Wikipedia & \num{2.68} & \num{2.70} & \num{93.4} / \num{92.0} \\
PubMed & \num{2.83} & \num{2.09} & \num{69.6} / \num{86.6} \\
Whole label & \num{1.05} & \num{0.88} & \num{79.5} / \num{90.7} \\
\bottomrule
\end{tabular}
\caption{Within-drug odds ratio when a fact counts as named, and when its sentence must also state the relation, as read by the model judge. Here we recompute named facts sentence by sentence; they agree with \cref{tab:coverage,tab:label} on \num{97.9}\% to \num{100}\% of edges. The last column gives the share of named facts (\%) whose sentence states the relation.}
\label{tab:relcheck}
\end{table}

%% file: tables/tab_label.tex
\begin{table}[t]
\centering
\footnotesize
\setlength{\tabcolsep}{4pt}
\begin{tabular}{lcccc}
\toprule
Text & Ind. & Con. & OR$_{\mathrm{MH}}$ & Deg. \\
\midrule
Wikipedia & \num{22.25} & \num{10.95} & & \\
Wiki + PubMed & \num{30.51} & \num{17.23} & & \\
Label, own section & \num{19.07} & \num{2.61} & \num{9.85} & ind.\ \\
Label, whole & \num{23.73} & \num{22.88} & \num{1.05} & contra.\ \\
All three & \num{34.04} & \num{26.69} & \num{1.43} & none \\
\bottomrule
\end{tabular}
\caption{Share of facts (\%) named on the same \num{1416} edges per relation (\num{472} drugs with a label). \emph{Deg.} is the relation that text favors once the drug's edge count is fixed in the logistic model: \emph{ind.} and \emph{contra.} are significant at $p$<\num{0.01}, and \emph{none} has $p$=\num{0.86}. Within-drug $p$-values: \num{e-45}, \num{0.62} and \num{2e-5}.}
\label{tab:label}
\end{table}

%% file: tables/tab_distractors.tex
\begin{table}[t]
\centering
\small
\setlength{\tabcolsep}{4pt}
\begin{tabular}{lccccc}
\toprule
Model & Ind. & Contra. & Gap & Within & $p$ \\
\midrule
Qwen2.5-0.5B & \num{36.0} & \num{28.1} & \num{+7.9} & \num{+8.2} & \num{0.009} \\
Qwen2.5-7B & \num{56.0} & \num{25.1} & \num{+30.9} & \num{+31.3} & \num{<e-15} \\
OLMo-2-7B & \num{56.0} & \num{33.0} & \num{+23.0} & \num{+23.6} & \num{2e-13} \\
Granite-3.1-8B & \num{58.7} & \num{12.8} & \num{+45.9} & \num{+46.3} & \num{<e-15} \\
\bottomrule
\end{tabular}
\caption{Closed-book accuracy (\%) with wrong options from the answer's MONDO neighborhood and a \emph{Not enough information} option; \num{450} indication and \num{406} contraindication questions. \emph{Within} is the gap computed within groups of questions whose page names the answer, its family or nothing. The $p$-value tests this within-group gap.}
\label{tab:distractors}
\end{table}

%% file: tables/tab_route.tex
\begin{table}[t]
\centering
\footnotesize
\setlength{\tabcolsep}{4.5pt}
\begin{tabular}{llcccc}
\toprule
Rel. & In label & $n$ & Prose & Label & Diff. \\
\midrule
Contra. & yes & \num{135} & \num{40.0} & \num{55.8} & \num{+15.8} \\
 & no & \num{211} & \num{17.1} & \num{12.2} & \num{-4.9} \\
\midrule
Ind. & yes & \num{117} & \num{84.3} & \num{87.7} & \num{+3.4} \\
 & no & \num{263} & \num{44.9} & \num{33.8} & \num{-11.0} \\
\bottomrule
\end{tabular}
\caption{Accuracy (\%) with a prose page and with a whole-label page, pooled over the three \num{7}B to \num{8}B models, split by whether the whole label names the answer (\emph{In label}).}
\label{tab:route}
\end{table}

%% file: Sections/6.dislimcon.tex
\section{Discussion}

\begin{figure*}[t]
\centering
\includegraphics[width=\textwidth]{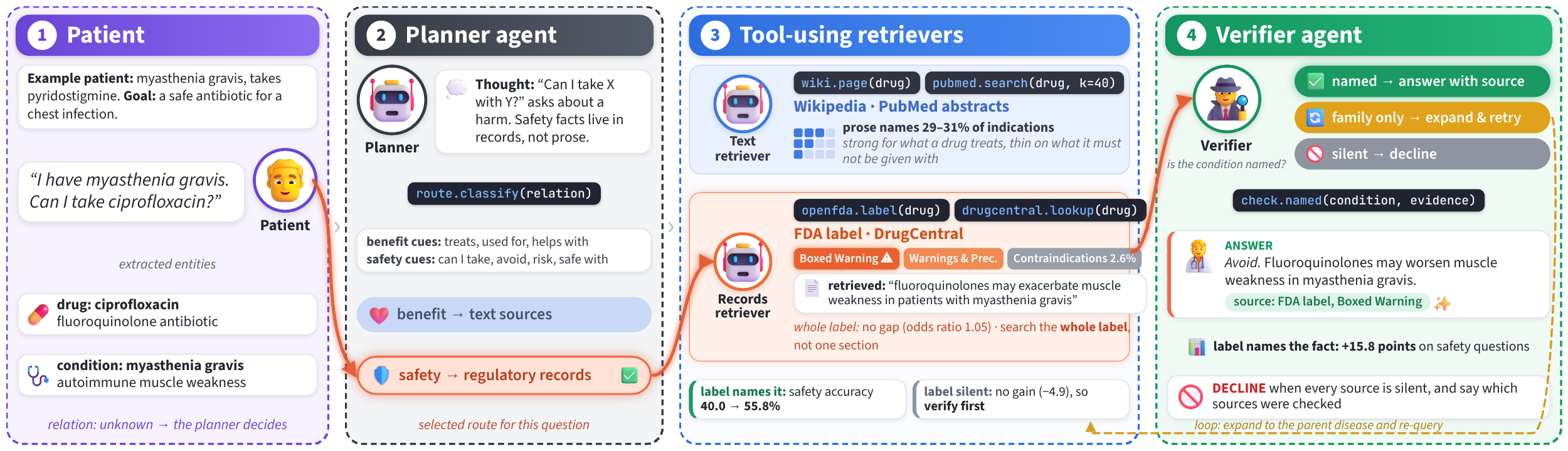}
\caption{A relation-aware clinical agent workflow that follows from our findings. (\num{1}) A planner classifies whether the question asks about a drug's benefits (\emph{indications}) or its safety limits (\emph{contraindications}), and routes it to general prose for benefits, or to structured databases and the whole FDA label for safety limits. (\num{2}) A verifier checks whether the retrieved passage names the exact condition, names only its family, or is silent, and the agent then answers with a citation, expands the family through the ontology, or abstains.}
\label{fig:router}
\end{figure*}

\paragraph{Medication safety is the blind spot of general prose.}
Two independent prose corpora, Wikipedia articles written for the public and PubMed abstracts written by researchers, show the same bias: the facts that matter most for preventing harm are named about half as often as the facts about what a drug treats. FDA labels, which must document both efficacy and safety, show no such gap. One explanation is that prose is written around a drug's purpose. An encyclopedia article introduces a drug by what it treats, and most clinical papers study whether a therapy works. Even papers on toxicity rarely list the preexisting conditions that make a drug unsafe. Scale does not fill this gap: across a \num{28}-fold increase in size, contraindication accuracy stays near chance, and larger models change how they fail, from wrong guesses to abstention, rather than whether they know (\cref{fig:failure}).

\paragraph{Design principles for clinical AI agents.}
Any clinical agent that answers medication questions from its own weights or from general prose inherits the \lbl{} asymmetry. \Cref{fig:router} shows a relation-aware design that follows from our results, of which we test the routing step.
\begin{enumerate}
\item \textbf{Relation-aware routing.} A planner first decides whether the question is about benefit or about safety. Benefit questions can go to general prose, which names \num{29}\% to \num{31}\% of indications. Safety questions should go to structured drug databases such as DrugCentral and to the \emph{whole} FDA label. The label's \emph{Contraindications} section alone names only \num{2.6}\% of PrimeKG's contraindications, against \num{22.9}\% for the whole label.
\item \textbf{Verification before answering.} A lightweight verifier checks how the retrieved text names the condition. Our string and ontology matcher is fast enough for this check. The routing test shows why it matters: the label passage raises contraindication accuracy by \num{+15.8} when it names the condition, and does not help when it is silent (\cref{tab:route}).
\item \textbf{Handling coarse or silent evidence.} With an exact or alternative-name match, the agent answers and cites the sentence. With only a family mention, it treats the passage as insufficient and uses the ontology to check specific subtypes. With silent text, it abstains and tells the user which source it checked.
\end{enumerate}

\section{Limitations}
\label{sec:limits}

\paragraph{Corpora and model weights.}
Model developers do not release their exact pretraining corpora, so the link between prose coverage and closed-book accuracy is a correlation, not a trace of pretraining tokens. The same gap in two prose sources, its absence in FDA labels, and model accuracy that follows prose coverage across three relations make pretraining exposure a plausible mechanism, but we do not establish it. Our PubMed analysis uses abstracts only, so its coverage is a lower bound for the full literature.

\paragraph{String matching and paraphrase.}
String matching miss clinical paraphrases and match a disease mentioned in passing. The random floor bounds false positives at under \num{1.3}\% in prose and \num{2.9}\% in FDA labels. The LLM judge, validated by annotators, finds that \num{87}\% to \num{93}\% of named contraindications and \num{70}\% to \num{93}\% of named indications state the relation.

\paragraph{Models and question format.}
The scaling curve uses one model family (Qwen2.5). The harder-option and open-book tests add two more families (OLMo-2 and Granite-3.1) at the \num{7}B to \num{8}B scale. Multiple choice measures whether a model recognizes the correct disease, which is an upper bound on its ability to recall it in free-form answers.

\paragraph{Granularity of graph facts.}
Even the whole FDA label names only \num{22.88}\% of PrimeKG's sampled contraindications, because PrimeKG often records a specific subtype (\emph{adult-onset myasthenia gravis}) while text states the parent condition (\emph{myasthenia gravis}). A clinical review of unmatched facts would separate such granularity differences from true omissions.

\section{Conclusion and Future Work}

General prose names a drug's benefits about twice as often as its safety limits, the FDA label does not. And LLMs follow the former. Several directions follow from this result. First, OLMo-2 releases its pretraining data, so the link between corpus and model can be traced directly by counting how often each fact appears in the tokens a model was trained on. Second, the audit can extend to full-text articles, clinical guidelines and other languages, and to safety relations beyond contraindications, such as drug interactions. Third, free-form answers and medical LLMs would show whether the gap holds beyond multiple choice and general-purpose models. Fourth, a clinical review of facts that no text names would separate graph granularity from true omission. Finally, we plan to build and evaluate the full relation-aware agent of \cref{fig:router} end to end, with a verifier that checks every safety answer against the label before the agent responds.

%% file: Appendix/0_contents.tex
\section*{Supplementary Material}
\addcontentsline{toc}{section}{Supplementary Material}

This supplement gives every detail needed to read, check and rebuild the results in the main paper. It holds the full data counts, the exact search queries and prompts, the matching rule and its pseudocode, every statistic, every table behind each figure, the annotation study, the routing test, the negative control and worked case studies. Each appendix stands on its own, so a reader can go straight to the part they need.

\paragraph{Contents.}
\begin{itemize}
\item \textbf{\Cref{app:overview}: Overview and reproducibility.} The workflow of the study as pseudocode, the pipeline stage by stage with its scripts, data versions and checksums, compute, the commands that rebuild every table, and a summary of every check we run against the main finding.
\item \textbf{\Cref{app:data}: Data and samples.} Counts for PrimeKG, Hetionet, MONDO and the Disease Ontology, how we build the disease hierarchy, how we draw each sample, and how plainly text names the facts in each sample.
\item \textbf{\Cref{app:corpora}: Corpora.} How we fetch Wikipedia, the exact PubMed queries, the single and balanced PubMed versions, and how we find and split FDA labels.
\item \textbf{\Cref{app:matching}: Measuring whether text names a fact.} The three sets of names, the matching rule with pseudocode, the chance floor, the floor-corrected lift of each naming level, and examples of facts found only at the coarser levels.
\item \textbf{\Cref{app:stats}: Statistical procedures.} Definitions of the Mantel-Haenszel odds ratio, its Robins-Breslow-Greenland interval, the edge-count control, Wilson intervals, the paired bootstrap, the within-group gap, the page effect and the agreement statistics.
\item \textbf{\Cref{app:withindrug}: Within-drug tests in every corpus.} The odds ratio with both kinds of interval, the logistic edge-count control and the edge-count strata for every corpus, including the first single-query PubMed corpus.
\item \textbf{\Cref{app:label}: FDA label analysis.} Coverage with intervals for every corpus and label section, what the label adds to prose, what the Contraindications section holds, and the label's chance floor.
\item \textbf{\Cref{app:probe}: Closed-book probe.} The prompt stems, how we build the options, the models, accuracy at every size with intervals, the paired gain from scale, and accuracy split by whether text names the fact.
\item \textbf{\Cref{app:page}: Harder wrong options and the drug's page.} The question format with a worked example, how we pick wrong options, how we cut the page, letter scoring with pseudocode, and every per-model table for the closed-book and page conditions.
\item \textbf{\Cref{app:relcheck}: Does the sentence state the relation?} The model judge prompt, the keyword baseline, the four-annotator study, agreement, the full table of definitions and the recall check on the label.
\item \textbf{\Cref{app:routing}: Routing test.} The setup and pseudocode of the routing step, per-model outcomes, and the split by whether the label names the answer.
\item \textbf{\Cref{app:hetionet}: Negative control on Hetionet.} The full control with its strata and naming levels.
\item \textbf{\Cref{app:cases}: Case studies.} One drug traced through every stage, safety facts that only the label names, questions that routing fixes, and the errors and edge cases we found.
\end{itemize}

\paragraph{How to read the tables.}
Unless a caption says otherwise, a percentage is the share of edges or questions, a difference is in percentage points, brackets hold a \num{95}\% interval, and \emph{Ind.} and \emph{Contra.} stand for indication and contraindication. A \emph{gap} is always indication minus contraindication, so a positive gap means text or models do better on benefit facts.

\paragraph{A note on terms.}
An \emph{edge} is one drug-disease fact in a knowledge graph. A fact is \emph{named} when the drug's text contains the disease's exact name or an alternative name (levels \num{0} and \num{1} in \cref{tab:levels}). \emph{Family only} means the text names only a broader MONDO term. \emph{Silent} means it names neither. A model \emph{declines} when it picks the option \emph{Not enough information to tell}. The code numbers the levels \num{0} (exact), \num{2} (alternative name), \num{3} (family) and \num{4} (none), and keeps level \num{1} for a synonym list that we do not use, so these four code levels map to levels \num{0}, \num{1}, \num{2} and \num{3} of the paper.

%% file: Appendix/A_overview.tex
\section{Overview and Reproducibility}
\label{app:overview}

This appendix gives the workflow of the whole study, the scripts behind each stage, the data versions, and the commands that rebuild every number. It ends with a summary of every check we run against the main finding.

\subsection{Workflow}

The study has one idea: hold the drug fixed and compare its two kinds of fact. The pseudocode below shows the whole workflow. Each step is a function described in a later appendix.

{\scriptsize
\begin{verbatim}
# 1. Facts (Appendix B)
G   = PrimeKG drug-disease edges
H   = MONDO hierarchy, broad terms removed
S   = 3 indications + 3 contraindications
      for every drug that has 3 of each

# 2. Text (Appendices C, D)
for drug d in S:
    T["wiki"]  = wikipedia(d)
    T["pm"]    = pubmed(d, "treat", 20)
               + pubmed(d, "harm", 20)
    T["label"] = openfda(d, max_labels=5)
    for edge (d, r, y) in S[d]:
        for c in T:
            level[c] = name_level(T[c], y, H)
            floor[c] = mean(name_level(T[c], y2, H)
                            for 3 unrelated y2)

# 3. Within-drug test (Appendices E, F, G)
for corpus c:
    OR, CI = mantel_haenszel(level[c] <= 1,
                             strata = drugs)
    coef   = logistic(named ~ relation
                      + log(edges of d in r))

# 4. Models (Appendices H, I, K)
for model m, edge e:
    closed[m, e] = answer(m, question(e))
    prose[m, e]  = answer(m, page_prose(e)
                              + question(e))
    label[m, e]  = answer(m, page_label(e)
                              + question(e))
compare by relation and by naming group

# 5. Checks (Appendices J, L)
judge(sentence) states the relation?
same test on Hetionet: two benefit relations
\end{verbatim}
}

\subsection{Pipeline}

The analysis runs as six stages. Each stage is a shell function in \texttt{pipeline.sh} that calls standalone Python scripts and reads only the files of the stages before it. \Cref{tab:stages} lists the stages and their scripts. One command, \texttt{bash run\_all.sh}, runs all six from the public downloads, submits the model jobs to the cluster scheduler, waits for them and writes every table. A second script, \texttt{relcheck.sh}, runs the relation check, scores the annotation sheets and runs the routing test.

\begin{table}[h]
\centering
\footnotesize
\setlength{\tabcolsep}{2.5pt}
\begin{tabular}{>{\raggedright\arraybackslash}p{1.3cm}>{\raggedright\arraybackslash}p{3.0cm}>{\raggedright\arraybackslash}p{3.45cm}}
\toprule
Stage & Scripts & What it does \\
\midrule
download & - & Downloads PrimeKG, MONDO, the Disease Ontology and Hetionet, and records their checksums. \\
prepare & prep\_primekg, mondo\_parents, prep\_hetionet & Splits the graphs by relation and builds the disease hierarchy tables. \\
text & gonogo\_p1, drug\_matched\_sample, second\_corpus, retier, name\_tiers & Draws the samples, fetches Wikipedia and PubMed, and assigns every edge a naming level and a floor. \\
extend & fetch\_labels, label\_coverage & Fetches FDA labels and measures naming in each label section. \\
models & run\_probe\_cpu, abstain\_probe & Runs the closed-book probe and the harder-option and page tests as cluster jobs. \\
analysis & dilution\_test, mh\_rbg, w3\_stats, w3\_extras, w3\_failure, w3\_summary & Computes every odds ratio, strata table, paired gain and page effect. \\
\midrule
relcheck & relcheck, route\_report & Runs the judge, the keyword check, the annotation scoring and the routing test. \\
\bottomrule
\end{tabular}
\caption{Pipeline stages and the scripts each stage runs.}
\label{tab:stages}
\end{table}

\paragraph{Caching.}
Every network request is cached on disk. A rerun reads the cache and makes no request. A failed request is never cached, so a network error is never stored as an empty result. Each stage writes a log with its request count, throttle events and unrecoverable failures. Every stage in our run reports zero unrecoverable failures.

\paragraph{Commands.}
The full run uses these commands. The smoke test runs every stage on a few edges in a few minutes, and the check script compares the rebuilt tables against the released ones.

{\scriptsize
\begin{verbatim}
bash run_all.sh --smoke     # every stage, tiny
bash run_all.sh             # full run
bash run_all.sh --status    # job progress
bash tools/check_results.sh # compare tables
bash relcheck.sh quick      # keywords, RBG CIs
bash relcheck.sh submit     # judge, label pages
bash relcheck.sh report     # judge, routing
bash relcheck.sh audit AUDIT.csv RECALL.csv
\end{verbatim}
}

\subsection{Data Versions and Compute}

\Cref{tab:versions} gives the files of our run with the first characters of their SHA-256 checksums. We downloaded all files on \num{2026}-\num{10}-\num{28}. Wikipedia, PubMed and openFDA change over time, so we release our cached copies of every article, abstract and label.

\begin{table}[h]
\centering
\footnotesize
\setlength{\tabcolsep}{3pt}
\begin{tabular}{lll}
\toprule
File & Version & SHA-256 (start) \\
\midrule
PrimeKG \texttt{kg.csv} & Dataverse & \texttt{0a5a010fcd88} \\
MONDO \texttt{mondo.obo} & 2026-10-06 & \texttt{2631e8e165d4} \\
DOID \texttt{doid.obo} & 2026-09-30 & \texttt{51e717b6b9f5} \\
Hetionet edges & v1.0 & \texttt{611f0411ac66} \\
Hetionet nodes & v1.0 & \texttt{3ad9eb86c031} \\
\bottomrule
\end{tabular}
\caption{Data files of our run and the start of their checksums.}
\label{tab:versions}
\end{table}

\paragraph{Compute.}
All model scoring runs on CPU nodes of a university supercomputer, with PyTorch 2.14 and Transformers 5.20 in \num{32}-bit floating point. No GPU is needed. Each model job has its own time limit by size, and the analysis job starts only after every model job has finished. The statistics use only the Python standard library, and every random draw uses seed \num{0}.

\paragraph{Released files.}
PrimeKG, Hetionet, MONDO, Wikipedia, PubMed and openFDA are public. We release the sampled edge lists, the naming level of every edge in every corpus, every model score, the annotation sheets and the scripts, so every table in the paper and in this supplement can be rebuilt from the released files without any network access.

\subsection{Summary of Checks}

\Cref{tab:checks} lists the alternative explanations for the gap and the tests that address them. Each row points to the appendix with the full result. The FDA label rules out the explanation that the facts are too rare to write down in any source.

\begin{table}[h]
\centering
\footnotesize
\setlength{\tabcolsep}{2.5pt}
\begin{tabular}{>{\raggedright\arraybackslash}p{2.5cm}>{\raggedright\arraybackslash}p{2.75cm}>{\raggedright\arraybackslash}p{2.35cm}}
\toprule
Explanation & Test & Result \\
\midrule
Famous drugs dominate & Within-drug odds ratio (\cref{app:withindrug}) & \num{2.19}-\num{2.77} in prose \\
Many contraindications per drug & Logistic model, edge-count strata (\cref{app:withindrug}) & Gap holds, $p\le\num{e-3}$ \\
One editorial process & Wikipedia and PubMed separately (\cref{app:withindrug}) & \num{2.43} and \num{2.77} \\
Retrieval favors efficacy & Split PubMed budget (\cref{app:corpora}) & OR \num{2.77} both ways \\
Matcher favors a relation & Hetionet, two benefit relations (\cref{app:hetionet}) & No gap, $p=\num{0.16}$ \\
Mention without relation & Judge, four annotators (\cref{app:relcheck}) & Prose \num{2.09}-\num{2.70}, label \num{0.88} \\
Matcher misses wording & Recall check, \num{50} facts (\cref{app:relcheck}) & \num{6}\% of misses \\
Coincidental matches & Floor on unrelated diseases (\cref{app:matching}) & $\le\num{1.2}$\% prose, $\le\num{2.8}$\% label \\
Facts too rare to write & Whole FDA label (\cref{app:label}) & OR \num{1.05}, $p=\num{0.62}$ \\
Small models only & Five sizes, 0.5B-14B (\cref{app:probe}) & Contra.\ \num{22}-\num{27}\% at every size \\
Easy wrong options & MONDO siblings, \num{4} models (\cref{app:page}) & Gap \num{7.9}-\num{45.9} points \\
Models cannot use safety text & Page in the prompt (\cref{app:page}) & \num{+14.2} vs.\ \num{+11.1} points \\
Label pages help everywhere & Routing split (\cref{app:routing}) & Only when the label names it \\
\bottomrule
\end{tabular}
\caption{Alternative explanations for the gap, the tests that address them and where to find each result.}
\label{tab:checks}
\end{table}

%% file: Appendix/B_data.tex
\section{Data and Samples}
\label{app:data}

This appendix gives the counts behind \cref{tab:data}, explains how we build the disease hierarchy, and shows how each sample is drawn.

\subsection{Knowledge Graphs}

\paragraph{PrimeKG.}
PrimeKG \citep{chandak2023primekg} ships as one edge file of \num{8100498} rows. We keep its three drug-disease relations and its disease-disease links (\cref{tab:primekg}). The drug-disease relations come from DrugCentral \citep{ursu2017drugcentral}. A drug has, on average, about three times as many contraindication edges as indication edges, which is why we test the edge count directly (\cref{app:withindrug}).

\begin{table}[h]
\centering
\footnotesize
\begin{tabular}{lr}
\toprule
PrimeKG relation & Edges \\
\midrule
indication & \num{9388} \\
contraindication & \num{30675} \\
off-label use & \num{2568} \\
disease-disease & \num{64388} \\
\midrule
all rows in \texttt{kg.csv} & \num{8100498} \\
\bottomrule
\end{tabular}
\caption{PrimeKG relations we use. The contraindication to indication ratio is \num{3.27}.}
\label{tab:primekg}
\end{table}

\paragraph{Hetionet.}
Hetionet v1.0 \citep{himmelstein2017hetionet} has \num{47031} nodes. Its \emph{treats} relation has \num{755} edges over \num{387} drugs and \num{77} diseases, and its \emph{palliates} relation has \num{390} edges over \num{221} drugs and \num{50} diseases. Both relations describe benefit, which makes Hetionet a negative control (\cref{app:hetionet}).

\subsection{Disease Hierarchy}

\paragraph{MONDO.}
We read the MONDO release of \num{2026}-\num{10}-\num{06}, which has \num{32134} terms and \num{46732} \emph{is a} links. We keep the links whose child carries a name in PrimeKG. Of the \num{17008} MONDO identifiers named in PrimeKG, \num{14591} (\num{85.8}\%) exist in this release. A family name is only useful if it is specific enough to point to a small group of diseases. We therefore drop every link whose parent has more than \num{400} descendants. This removes \num{5857} links and leaves \num{17801} links over \num{11788} diseases, with at most \num{9} parents per disease. \Cref{tab:broad} lists the broadest terms that the cut removes. Without the cut, the phrase ``disease'' would make almost every disease count as named at the family level.

\begin{table}[h]
\centering
\footnotesize
\setlength{\tabcolsep}{3pt}
\begin{tabular}{lr}
\toprule
MONDO term & Descendants \\
\midrule
disease or disorder & \num{31582} \\
human disease & \num{23262} \\
disease by body system or component & \num{18875} \\
disease by etiologic mechanism & \num{17810} \\
disease of genetic or genomic mechanism & \num{12256} \\
Mendelian disease & \num{11784} \\
non-human animal disease & \num{8318} \\
disease by developmental or physiological process & \num{7322} \\
\bottomrule
\end{tabular}
\caption{The broadest MONDO terms, all removed as family names by the cut at \num{400} descendants.}
\label{tab:broad}
\end{table}

\paragraph{Disease Ontology.}
Hetionet names diseases with the Disease Ontology \citep{schriml2022doid}, so we build its hierarchy the same way from the release of \num{2026}-\num{09}-\num{30}. It has \num{12335} terms. After the same cut, it gives \num{103} links over \num{74} of Hetionet's diseases. Examples are \emph{Crohn's disease} under \emph{inflammatory bowel disease} and \emph{Parkinson's disease} under \emph{synucleinopathy}.

\subsection{Samples}

\paragraph{Text sample.}
A within-drug comparison needs drugs that have both relations. We keep the \num{595} drugs with at least three indication and three contraindication edges and draw exactly three of each, so every drug contributes a balanced table. This gives \num{1785} edges per relation. Of these drugs, \num{574} have an English Wikipedia article long enough to measure, which leaves \num{1722} edges per relation. At this size, and at naming rates near \num{20}\% and \num{7}\%, the within-drug test detects an odds ratio of about \num{1.6} or larger. The \num{472} drugs with an FDA label give \num{1416} edges per relation.

\paragraph{Probe sample.}
The model probe uses a separate draw of \num{600} edges for each of indication, contraindication and off-label use, \num{1800} edges in all over \num{817} Wikipedia articles. This draw is not balanced within drugs. The probe compares models on the same edges, so the pairing comes from the edges themselves.

\paragraph{Harder-option sample.}
The harder-option test reuses the question set of our companion study on declining. It draws up to \num{150} questions for each relation and naming group, which gives \num{450} indication and \num{406} contraindication questions. Only \num{106} contraindications in the pool are named only at the family level, so that group is smaller.

\paragraph{Hetionet sample.}
We draw up to \num{600} edges per relation. \emph{Palliates} has only \num{390} edges, of which \num{385} have a measurable article, so the sample has \num{600} \emph{treats} and \num{385} \emph{palliates} edges.

\paragraph{How plainly the samples are named.}
\Cref{tab:sampletiers} gives the naming level of every edge in each sample, measured on Wikipedia. Most PrimeKG facts are silent, while most Hetionet facts are named. Hetionet's diseases are fewer and broader, so its articles name them more often. This difference does not matter for the control, because the control compares two relations inside Hetionet.

\begin{table}[h]
\centering
\footnotesize
\setlength{\tabcolsep}{3pt}
\begin{tabular}{lrcccc}
\toprule
Sample & Edges & Exact & Alt. & Family & None \\
\midrule
Text (PrimeKG) & \num{3444} & \num{13.2} & \num{1.8} & \num{9.8} & \num{75.1} \\
Probe (PrimeKG) & \num{1800} & \num{10.3} & \num{1.0} & \num{10.1} & \num{78.6} \\
Hetionet & \num{985} & \num{54.5} & \num{2.3} & \num{5.3} & \num{37.9} \\
\bottomrule
\end{tabular}
\caption{Naming level of every sampled edge in Wikipedia (\%), both relations together.}
\label{tab:sampletiers}
\end{table}

%% file: Appendix/C_corpora.tex
\section{Corpora}
\label{app:corpora}

This appendix explains how we fetch each corpus and how we check that retrieval does not create the gap.

\subsection{Wikipedia}

For each drug, we fetch the current English Wikipedia article by title through the MediaWiki API and cache its source text. We use the article as fetched, with its markup. The relation check removes markup before it splits sentences (\cref{app:relcheck}). The \num{574} drugs of the text sample have \num{574} articles, and the \num{1800} probe edges come from \num{817} articles.

\subsection{PubMed}

\paragraph{Queries.}
We retrieve abstracts through the NCBI E-utilities \citep{sayers2009eutils}, sorted by relevance. Each drug gets a budget of \num{40} abstracts, split evenly between two searches with \num{20} abstracts each. Both searches require the drug's name in the title or abstract. One adds a fragment about therapeutic use, and the other adds a fragment about harm:

{\scriptsize
\begin{verbatim}
"<drug>"[Title/Abstract] AND
  (treat*[tiab] OR therap*[tiab]
   OR efficac*[tiab])

"<drug>"[Title/Abstract] AND
  (contraindicat*[tiab] OR adverse[tiab]
   OR toxicit*[tiab] OR "side effect"[tiab]
   OR "side effects"[tiab] OR induced[tiab])
\end{verbatim}
}

We fetch the union of returned records once, join the titles and abstracts into one text per drug, and cache it. A text shorter than \num{400} characters counts as absent, which affects \num{12} of \num{3444} edges. The balanced run makes \num{1721} requests with no unrecoverable failure.

\paragraph{Why a balanced search.}
A search for a drug's name alone, sorted by relevance, could return mostly efficacy papers and so create the gap by retrieval. The first version of the corpus used exactly such a search, with \num{40} abstracts per drug. The balanced search gives the harm fragment the same budget as the treatment fragment. If retrieval created the gap, the balanced search would shrink it. It does not: both versions give a within-drug odds ratio of \num{2.77} (\cref{tab:or_all}).

\paragraph{The silent share.}
\Cref{tab:silent} gives the share of facts that a corpus does not name, under each version. Adding PubMed to Wikipedia shrinks the silent share by \num{7.40} points with the balanced search and \num{5.81} points with the single search. The silent share stays far higher for contraindications in every column.

\begin{table}[h]
\centering
\footnotesize
\setlength{\tabcolsep}{3pt}
\begin{tabular}{llccc}
\toprule
Search & Relation & Wikipedia & PubMed & Union \\
\midrule
Balanced & contraindication & \num{90.19} & \num{88.56} & \num{84.09} \\
 & indication & \num{79.67} & \num{73.69} & \num{70.96} \\
 & both & \num{84.93} & \num{81.13} & \num{77.53} \\
\midrule
Single & contraindication & \num{90.19} & \num{89.78} & \num{84.90} \\
 & indication & \num{79.67} & \num{76.66} & \num{73.34} \\
 & both & \num{84.93} & \num{83.22} & \num{79.12} \\
\bottomrule
\end{tabular}
\caption{Share of facts (\%) that a corpus does not name (level above \num{1}), \num{1722} edges per relation.}
\label{tab:silent}
\end{table}

\subsection{FDA Label}

\paragraph{Finding a label.}
We query the openFDA drug label service \citep{kasshout2016openfda} by generic name, then by substance name, then by brand name, and keep the first search that returns a label. A drug sold by several manufacturers has one label from each, so we pool up to five labels per drug. We find a label for \num{472} of the \num{574} drugs (\num{82.2}\%).

\paragraph{Sections.}
From each label we keep three texts. The \emph{Indications and Usage} section is the indication's own section. The \emph{Contraindications} section is the contraindication's own section. The \emph{whole label} joins the text of every section, including the Boxed Warning, Warnings and Precautions, Adverse Reactions and Use in Specific Populations. We measure indications and contraindications in their own section and in the whole label (\cref{app:label}).

\paragraph{Combination products.}
A search by generic name can return the label of a combination product. For example, a search for tretinoin can return a label for clindamycin and tretinoin, whose warnings about colitis come from clindamycin. We keep these labels, because a retriever would return them too. \Cref{app:cases} shows one such case.

%% file: Appendix/D_matching.tex
\section{Measuring Whether Text Names a Fact}
\label{app:matching}

This appendix explains how we decide whether a drug's text names a disease. The same rules apply to Wikipedia, PubMed and the FDA label.

\subsection{Names and Matching}

\paragraph{Three sets of names.}
Each disease has three sets of names. The \emph{exact name} is its MONDO label \citep{vasilevsky2022mondo}. The \emph{alternative names} are the titles of Wikipedia pages that redirect to the disease's article, which bring in synonyms, lay terms and abbreviations. For example, \emph{juvenile diabetes} redirects to \emph{type 1 diabetes}. The \emph{family names} are the labels of the disease's MONDO ancestors up to two levels above it, after the cut of \cref{app:data}. On the draw of \cref{tab:floor}, a disease has \num{1.5} to \num{2.1} family names on average, depending on the relation. The sets are nested: a text that names a disease at a finer level also counts at every coarser level, and each edge gets the finest level it reaches (\cref{tab:levels}).

\paragraph{Matching rule.}
A phrase matches when it appears on word boundaries, ignoring case, with an optional trailing ``s'' for plurals. A multi-word phrase matches across any run of white space, so a line break inside a name does not break the match. We never match phrases shorter than four characters, so short abbreviations do not create chance hits. Under this rule, ``asthma'' does not match ``asthmatic'', and ``colitis'' does not match inside a longer word.

{\scriptsize
\begin{verbatim}
def matches(text, names):
    for n in names:
        if len(n) < 4:
            continue
        words = n.split()
        pat = (r"\b" + r"\s+".join(
                   re.escape(w) for w in words)
               + r"s?\b")
        if re.search(pat, text, re.I):
            return True
    return False

def name_level(text, y, H):
    if matches(text, [label(y)]):     return 0
    if matches(text, redirects(y)):   return 1
    if matches(text, ancestors(y, H,
                       depth=2)):     return 2
    return 3

named = name_level(text, y, H) <= 1
\end{verbatim}
}

\paragraph{What counts as named.}
In the main paper, only levels \num{0} and \num{1} count as named. A family match does not count, because ``diabetes'' does not say which diabetes the drug treats or harms. We report the family level separately, because it shows where text points in the right direction without naming the fact.

\subsection{Chance Floor}

For every edge, we draw three diseases from the same relation's targets that have no edge to the drug in that relation. We run the same matcher on the drug's text for these unrelated diseases. The share that matches is the \emph{floor}. It estimates how often a disease name appears in a drug's text by coincidence, not as a statement of the relation. We compute the floor at every level, because wider name sets also match unrelated diseases more often.

{\scriptsize
\begin{verbatim}
def floor(edges, text, level):
    hits = 0
    for (d, r, y) in edges:
        pool = [y2 for y2 in targets(r)
                if (d, r, y2) not in G]
        for y2 in sample(pool, 3, seed=0):
            hits += name_level(text[d], y2, H)
                        <= level
    return hits / (3 * len(edges))
\end{verbatim}
}

\Cref{tab:floor} gives the real rate and the floor at each level on a draw of \num{600} edges per relation from the full graph. At the exact level, real edges match about \num{7} to \num{15} times more often than unrelated ones. At the family level, the floor rises to \num{1.83}-\num{3.28}\%, still far below the threshold of about \num{15}\% at which a level would only measure how common its words are.

\begin{table}[h]
\centering
\footnotesize
\setlength{\tabcolsep}{2.5pt}
\begin{tabular}{lcccccc}
\toprule
 & \multicolumn{3}{c}{Real edge} & \multicolumn{3}{c}{Floor} \\
\cmidrule(lr){2-4}\cmidrule(lr){5-7}
Relation & exact & +alt. & +family & exact & +alt. & +family \\
\midrule
Indication & \num{17.17} & \num{19.33} & \num{34.67} & \num{1.22} & \num{1.39} & \num{2.67} \\
Off-label & \num{10.50} & \num{11.83} & \num{23.67} & \num{0.72} & \num{0.94} & \num{3.28} \\
Contraind. & \num{5.50} & \num{6.67} & \num{9.33} & \num{0.78} & \num{0.89} & \num{1.83} \\
\bottomrule
\end{tabular}
\caption{Cumulative match rate (\%) in Wikipedia for real edges and for three unrelated diseases per edge, \num{600} edges per relation. Each column adds a set of names to the one before.}
\label{tab:floor}
\end{table}

\subsection{What Each Level Adds}

We split the lift of each level into two parts. \emph{Nomenclature} is the floor-corrected gain from alternative names: it measures how often text uses another word for the same disease. \emph{Granularity} is the floor-corrected gain from family names: it measures how often text names only a broader group. \Cref{tab:lift} gives both parts.

\begin{table}[h]
\centering
\footnotesize
\setlength{\tabcolsep}{3pt}
\begin{tabular}{lccc}
\toprule
Relation & Exact, corrected & Nomenclature & Granularity \\
\midrule
Indication & \num{15.94} & \num{2.00} & \num{14.06} \\
Off-label use & \num{9.78} & \num{1.11} & \num{9.50} \\
Contraindication & \num{4.72} & \num{1.06} & \num{1.72} \\
\midrule
Mean & & \num{1.39} & \num{8.43} \\
\bottomrule
\end{tabular}
\caption{Floor-corrected naming rate (\%) at the exact level, and the floor-corrected gain (points) from alternative names and from family names up to two levels.}
\label{tab:lift}
\end{table}

Three things stand out. First, alternative names add little for every relation, so the gap is not a matter of synonyms. Second, family names add a lot for benefit facts: text often says that a drug treats ``bacterial infections'' rather than a specific infection. Third, family names add almost nothing for contraindications. Text does not name most contraindications even at the level of their family, so no name table or ontology lookup can recover them from prose.

\paragraph{One level up.}
With only the immediate parent as the family name, the granularity gain is \num{9.33} points for indications, \num{5.83} for off-label uses and \num{1.33} for contraindications, with a mean of \num{5.50}. The order of the relations is the same at both depths.

\subsection{Examples}

\Cref{tab:recovered} lists contraindications that Wikipedia does not name exactly but names at a coarser level. These are the cases that the family level exists to catch. A family match says that the text names a broader disease that contains the fact, not the fact itself. An alternative-name match is a real naming, because a redirect is another name for the same disease.

\begin{table}[h]
\centering
\footnotesize
\setlength{\tabcolsep}{3pt}
\begin{tabular}{lll}
\toprule
Drug & Contraindicated disease & Found as \\
\midrule
Isoniazid & acute liver failure & family \\
Entecavir & non-alcoholic steatohepatitis & family \\
Ganciclovir & adult idiopathic neutropenia & family \\
Prednisolone & ocular tuberculosis & family \\
Clonidine & chronic renal failure syndrome & family \\
Sumatriptan & coronary thrombosis & family \\
Ciprofloxacin & Achilles bursitis & family \\
Salbutamol & hypertension & alt.\ name \\
Almotriptan & Prinzmetal angina & alt.\ name \\
Pazopanib & long QT syndrome & alt.\ name \\
Clomifene & vision disorder & alt.\ name \\
\bottomrule
\end{tabular}
\caption{Contraindications that Wikipedia names only by an alternative name or by their family.}
\label{tab:recovered}
\end{table}

%% file: Appendix/E_stats.tex
\section{Statistical Procedures}
\label{app:stats}

This appendix defines every statistic in the paper and in this supplement. The statistics use only the Python standard library, and every random draw uses seed \num{0}.

\paragraph{Within-drug odds ratio.}
For drug $d$, let $a_d$ and $b_d$ be its named and unnamed indication edges, $c_d$ and $e_d$ its named and unnamed contraindication edges, and $n_d$ their total. The Mantel-Haenszel estimate \citep{mantel1959} is
\begin{equation}
\mathrm{OR}_{\mathrm{MH}} = \frac{\sum_d a_d e_d / n_d}{\sum_d b_d c_d / n_d}.
\end{equation}
With $P_d=(a_d+e_d)/n_d$, $Q_d=(b_d+c_d)/n_d$, $R_d=a_d e_d/n_d$ and $S_d=b_d c_d/n_d$, we use the Robins-Breslow-Greenland variance \citep{robins1986},
\begin{align}
\widehat{\mathrm{Var}}\big(\log \mathrm{OR}_{\mathrm{MH}}\big) ={}& \frac{\sum_d P_d R_d}{2R^2} + \frac{\sum_d Q_d S_d}{2S^2} \nonumber\\
&+ \frac{\sum_d (P_d S_d + Q_d R_d)}{2RS},
\end{align}
where $R=\sum_d R_d$ and $S=\sum_d S_d$. The \num{95}\% interval is $\exp\big(\log \mathrm{OR}_{\mathrm{MH}} \pm \num{1.96}\,\widehat{\mathrm{SE}}\big)$. This variance stays valid when each drug contributes only a few edges, as ours do. We test the null hypothesis of no association with the continuity-corrected Cochran-Mantel-Haenszel test.

\paragraph{Edge-count control.}
A logistic model predicts whether a fact is named from an indicator for its relation and from the logarithm of the drug's number of edges in that relation in the full graph. The relation coefficient tests the gap with the edge count held fixed.

\paragraph{Wilson intervals.}
For $k$ successes in $n$ trials, with $\hat p = k/n$ and $z = \num{1.96}$, the interval is
\begin{equation}
\frac{\hat p + \frac{z^2}{2n} \pm z\sqrt{\frac{\hat p(1-\hat p)}{n} + \frac{z^2}{4n^2}}}{1 + \frac{z^2}{n}}.
\end{equation}
Unlike the simple normal interval, it stays inside $[0, 1]$ and behaves well for small shares such as the chance floor.

\paragraph{Paired gain from scale.}
For each edge $i$, the gain is $g_i = y_i^{\mathrm{7B}} - y_i^{\mathrm{0.5B}}$, where $y=1$ for a correct answer and $0$ otherwise. The statistic is the mean gain on indications minus the mean gain on contraindications. Its interval comes from \num{5000} bootstrap resamples in which we draw drugs, not edges, with replacement, so the edges of one drug move together. The $p$-value uses the normal approximation with the bootstrap standard error.

\paragraph{Gap within naming groups.}
In the harder-option test, we compute the indication-minus-contraindication difference $\delta_g$ separately within the questions whose page names the answer, names only its family, or names nothing. We combine the groups as $\sum_g w_g \delta_g / \sum_g w_g$ with $w_g = n_{g,1} n_{g,2} / (n_{g,1} + n_{g,2})$, where $n_{g,1}$ and $n_{g,2}$ are the group's indication and contraindication counts. The standard error follows from the binomial variances of the group proportions.

\paragraph{Page effect.}
For the page test, the effect of the page on a question is its outcome with the page minus its outcome closed book. We average this effect over questions and draw \num{95}\% intervals from \num{5000} bootstrap resamples of questions. For pooled results, we first average each question's effect over the three 7-8B models, so each question counts once.

\paragraph{Agreement between annotators.}
For $N$ items, each labeled by $n$ annotators, let $n_{ij}$ be the number of annotators who give item $i$ label $j$. Fleiss' kappa is $\kappa = (\bar P - \bar P_e)/(1 - \bar P_e)$, where $\bar P = \frac{1}{N}\sum_i \frac{\sum_j n_{ij}^2 - n}{n(n-1)}$ is the observed agreement and $\bar P_e = \sum_j p_j^2$, with $p_j$ the overall share of label $j$, is the agreement expected by chance. Cohen's kappa is the same quantity for two raters.

\paragraph{Bootstrap over drugs.}
The pseudocode below shows the paired bootstrap of the gain from scale. Resampling drugs rather than edges keeps the edges of one drug together, so a drug with many edges cannot make the interval look narrower than it is.

{\scriptsize
\begin{verbatim}
def boot_gain(edges, B=5000, seed=0):
    by_drug = group(edges, key="drug")
    drugs = list(by_drug)
    stats = []
    for b in range(B):
        pick = choices(drugs, k=len(drugs))
        E = [e for d in pick for e in by_drug[d]]
        g_ind = mean(e.y7 - e.y05 for e in E
                     if e.rel == "indication")
        g_con = mean(e.y7 - e.y05 for e in E
                     if e.rel == "contraindication")
        stats.append(g_ind - g_con)
    return percentile(stats, [2.5, 97.5])
\end{verbatim}
}

\paragraph{Why not compare per-drug rates.}
A paired test of per-drug naming rates would look natural, but it is not valid here. With three edges per relation, a drug's rate can only be \num{0}, \num{1}/\num{3}, \num{2}/\num{3} or \num{1}, so the comparison would measure this rounding as much as the gap. The Mantel-Haenszel estimate uses the counts directly and has no such problem.

%% file: Appendix/F_withindrug.tex
\section{Within-Drug Tests in Every Corpus}
\label{app:withindrug}

This appendix gives the full within-drug results for every corpus: the raw naming rates, the Mantel-Haenszel odds ratio with both kinds of interval, the logistic edge-count control and the edge-count strata.

\subsection{Odds Ratios}

\Cref{tab:or_all} gives the naming rates and the within-drug odds ratio for every corpus. The test-based interval comes from the Cochran-Mantel-Haenszel statistic. The Robins-Breslow-Greenland (RBG) interval comes from the variance of \cref{app:stats}. The two intervals agree closely in every corpus, so the result does not depend on the choice. Both prose corpora show a gap on their own, and so does each PubMed version. The whole label shows none.

\begin{table}[h]
\centering
\scriptsize
\setlength{\tabcolsep}{1.6pt}
\begin{tabular}{lcccc}
\toprule
Corpus & Ind. & Contra. & OR [RBG \num{95}\% CI] & CMH $p$ \\
\midrule
Wikipedia & \num{20.33} & \num{9.81} & \num{2.43} [\num{1.98}, \num{2.98}] & \num{1.9e-18} \\
PubMed, balanced & \num{26.31} & \num{11.44} & \num{2.77} [\num{2.30}, \num{3.33}] & \num{5.5e-29} \\
PubMed, sgl. & \num{23.34} & \num{10.22} & \num{2.77} [\num{2.27}, \num{3.37}] & \num{1.9e-25} \\
Wiki + PubMed & \num{29.04} & \num{15.91} & \num{2.19} [\num{1.85}, \num{2.59}] & \num{1.7e-20} \\
Wiki + PubMed, sgl. & \num{26.66} & \num{15.10} & \num{2.10} [\num{1.76}, \num{2.49}] & \num{3.0e-17} \\
\midrule
Label, own section & \num{19.07} & \num{2.61} & \num{9.85} [\num{6.71}, \num{14.45}] & \num{1.3e-45} \\
Label, whole & \num{23.73} & \num{22.88} & \num{1.05} [\num{0.88}, \num{1.25}] & \num{0.62} \\
All three & \num{34.04} & \num{26.69} & \num{1.43} [\num{1.21}, \num{1.68}] & \num{2.2e-5} \\
\midrule
Hetionet & \num{57.33} & \num{56.10} & \num{2.48} [\num{1.21}, \num{5.08}] & \num{0.017} \\
\bottomrule
\end{tabular}
\caption{Share of facts named (\%) and the within-drug odds ratio. \emph{Sgl.} is the first, single-query PubMed corpus. Prose rows use \num{1722} edges per relation over \num{574} drugs, label rows \num{1416} edges over \num{472} drugs. For Hetionet, the columns are \emph{treats} and \emph{palliates}, and the odds ratio uses the \num{51} drugs that have both. Test-based intervals: Wikipedia [\num{2.01}, \num{2.94}], PubMed [\num{2.33}, \num{3.28}], PubMed single [\num{2.31}, \num{3.32}], Wiki + PubMed [\num{1.86}, \num{2.57}], Wiki + PubMed single [\num{1.77}, \num{2.48}], label own [\num{7.75}, \num{12.51}], label whole [\num{0.88}, \num{1.25}], all three [\num{1.21}, \num{1.68}], Hetionet [\num{1.29}, \num{4.75}].}
\label{tab:or_all}
\end{table}

\paragraph{Reading the label rows.}
The label's own sections show a very large gap, but this is a property of the sections, not of the label. The Indications and Usage section is written to list what a drug treats. The Contraindications section lists only formal contraindications, which are a small part of what PrimeKG calls a contraindication (\cref{app:relcheck}). Most safety facts in the label sit in its warnings and adverse reactions. The whole label, which holds all of them, is the fair comparison with prose.

\paragraph{Why the Hetionet odds ratio is above one.}
The Hetionet within-drug odds ratio is \num{2.48}, but it rests on only \num{51} drugs, and it does not survive the edge-count control below. \Cref{app:hetionet} explains this case.

\subsection{Edge-Count Control}

A drug with many contraindications may name each one less often, simply because no text lists them all. To test this, a logistic model predicts whether a fact is named from an indicator for contraindication and from the logarithm of the drug's number of edges in that relation in the full graph. \Cref{tab:logit} gives the coefficients. A negative relation coefficient means that a contraindication is less likely to be named than an indication of a drug with the same edge count.

\begin{table}[h]
\centering
\scriptsize
\setlength{\tabcolsep}{2pt}
\begin{tabular}{lcccc}
\toprule
Corpus & Contra.\ coef. & $p$ & log edges & $p$ \\
\midrule
Wikipedia & \num{-0.434} & \num{2.4e-4} & \num{-0.413} & \num{1.2e-10} \\
PubMed, balanced & \num{-0.473} & \num{1.4e-5} & \num{-0.554} & \num{1.2e-19} \\
PubMed, sgl. & \num{-0.472} & \num{3.5e-5} & \num{-0.517} & \num{3.6e-16} \\
Wiki + PubMed & \num{-0.329} & \num{9.7e-4} & \num{-0.437} & \num{8.2e-16} \\
\midrule
Label, own section & \num{-1.926} & \num{6.0e-22} & \num{-0.230} & \num{6.9e-3} \\
Label, whole & \num{+0.304} & \num{5.0e-3} & \num{-0.320} & \num{1.5e-8} \\
All three & \num{+0.017} & \num{0.86} & \num{-0.335} & \num{1.2e-10} \\
\midrule
Hetionet & \num{-0.191} & \num{0.16} & \num{-0.447} & \num{3.5e-9} \\
\bottomrule
\end{tabular}
\caption{Logistic model of whether a fact is named, with the relation and the log of the drug's edge count as predictors. For Hetionet, the relation term is \emph{palliates}.}
\label{tab:logit}
\end{table}

In every prose corpus, both terms matter. Drugs with more edges have each edge named less often, and contraindications are named less often even with the edge count held fixed. In the whole label, the sign flips: with the edge count held fixed, the label names a contraindication slightly more often than an indication. Once prose and label are pooled, the relation term vanishes.

\subsection{Edge-Count Strata}

\Cref{tab:strata} gives the gap within three strata of the drug's edge count in the relation. In prose, the gap is largest for drugs with few edges and shrinks as drugs have more. For drugs with \num{13} or more edges, it is not distinguishable from zero in any prose corpus. These drugs have mostly contraindications and few indications, and their texts name few facts of either kind. In the whole label, the gap is close to zero or negative in every stratum.

\begin{table}[h]
\centering
\footnotesize
\setlength{\tabcolsep}{2.5pt}
\begin{tabular}{lccc}
\toprule
Edges per drug & \num{3}-\num{5} & \num{6}-\num{12} & \num{13}+ \\
\midrule
Ind.\ edges ($n$, prose) & \num{756} & \num{678} & \num{288} \\
Contra.\ edges ($n$, prose) & \num{132} & \num{405} & \num{1185} \\
\midrule
Wikipedia & \num{+13.94}$^{***}$ & \num{+6.04}$^{**}$ & \num{+1.64} \\
PubMed, balanced & \num{+18.33}$^{***}$ & \num{+7.71}$^{***}$ & \num{+2.37} \\
PubMed, sgl. & \num{+17.88}$^{***}$ & \num{+7.57}$^{***}$ & \num{+1.92} \\
Wiki + PubMed & \num{+11.81}$^{**}$ & \num{+6.90}$^{**}$ & \num{+2.23} \\
\midrule
Label, own section & \num{+20.77}$^{***}$ & \num{+14.57}$^{***}$ & \num{+10.10}$^{***}$ \\
Label, whole & \num{-3.59} & \num{-6.46}$^{*}$ & \num{-1.50} \\
All three & \num{+5.14} & \num{-2.54} & \num{+1.88} \\
\bottomrule
\end{tabular}
\caption{Indication minus contraindication naming rate (points) within strata of the drug's edge count. $^{*}p<\num{0.05}$, $^{**}p<\num{0.01}$, $^{***}p<\num{0.001}$. Label strata have \num{594}, \num{591} and \num{231} indication edges and \num{102}, \num{297} and \num{1017} contraindication edges.}
\label{tab:strata}
\end{table}

\Cref{tab:strata_rates} gives the rates behind the Wikipedia and whole-label rows, so the reader can see that the shrinking gap in prose comes from indications, whose naming rate falls with edge count, while contraindications stay low in every stratum.

\begin{table}[h]
\centering
\footnotesize
\setlength{\tabcolsep}{3pt}
\begin{tabular}{llccc}
\toprule
Corpus & Relation & \num{3}-\num{5} & \num{6}-\num{12} & \num{13}+ \\
\midrule
Wikipedia & Ind. & \num{26.06} & \num{18.14} & \num{10.42} \\
 & Contra. & \num{12.12} & \num{12.10} & \num{8.78} \\
PubMed & Ind. & \num{36.51} & \num{20.80} & \num{12.50} \\
 & Contra. & \num{18.18} & \num{13.09} & \num{10.13} \\
Label, whole & Ind. & \num{27.78} & \num{21.49} & \num{19.05} \\
 & Contra. & \num{31.37} & \num{27.95} & \num{20.55} \\
\bottomrule
\end{tabular}
\caption{Naming rate (\%) by relation within strata of the drug's edge count.}
\label{tab:strata_rates}
\end{table}

%% file: Appendix/G_label.tex
\section{FDA Label Analysis}
\label{app:label}

The FDA label is the one corpus that regulation requires to carry a drug's safety facts. This appendix gives the full label results on the \num{472} drugs that have a label, with \num{1416} edges per relation.

\subsection{Coverage}

\Cref{tab:label_full} gives the share of facts named by each corpus and label section, with Wilson intervals. Every column uses the same drugs and the same edges, so the columns can be compared directly.

\begin{table}[h]
\centering
\footnotesize
\setlength{\tabcolsep}{3pt}
\begin{tabular}{lcc}
\toprule
Corpus & Indication & Contraindication \\
\midrule
Wikipedia & \num{22.25} [\num{20.2}, \num{24.5}] & \num{10.95} [\num{9.4}, \num{12.7}] \\
Wiki + PubMed & \num{30.51} [\num{28.2}, \num{33.0}] & \num{17.23} [\num{15.4}, \num{19.3}] \\
Label, own section & \num{19.07} [\num{17.1}, \num{21.2}] & \num{2.61} [\num{1.9}, \num{3.6}] \\
Label, whole & \num{23.73} [\num{21.6}, \num{26.0}] & \num{22.88} [\num{20.8}, \num{25.1}] \\
All three & \num{34.04} [\num{31.6}, \num{36.5}] & \num{26.69} [\num{24.5}, \num{29.1}] \\
\midrule
Floor, whole label & \num{2.10} & \num{2.82} \\
\bottomrule
\end{tabular}
\caption{Share of facts named (\%) with \num{95}\% Wilson intervals, \num{1416} edges per relation over \num{472} drugs with a label. The floor is the exact-name match rate of three unrelated diseases per edge.}
\label{tab:label_full}
\end{table}

Three results stand out. First, the whole label names contraindications as often as indications, at about \num{23}\% each, while Wikipedia names contraindications at half the rate of indications. Second, the whole label alone names more contraindications than Wikipedia and PubMed together. Third, the Contraindications section alone names only \num{2.61}\% of PrimeKG's contraindications. A retriever that indexes only this section would miss almost every safety fact that the label carries.

\subsection{What the Label Adds to Prose}

\Cref{tab:rescue} looks at the facts that prose misses, that is, facts that neither Wikipedia nor PubMed names. It asks how many of them the label names. The label rescues more than twice as many missed contraindications as missed indications, so it fills the gap that prose leaves.

\begin{table}[h]
\centering
\footnotesize
\setlength{\tabcolsep}{3pt}
\begin{tabular}{lccc}
\toprule
Relation & Missed by prose & Own section & Whole label \\
\midrule
Indication & \num{984} & \num{3.4} & \num{5.1} \\
Contraindication & \num{1172} & \num{1.4} & \num{11.4} \\
\bottomrule
\end{tabular}
\caption{Facts that prose misses (count) and the share of them (\%) that the label names, in the relation's own section or anywhere in the label.}
\label{tab:rescue}
\end{table}

\subsection{What the Contraindications Section Holds}

The Contraindications section is short and formal. It is empty for \num{9.5}\% of the contraindication edges, because some labels list no contraindications at all, while the Indications section is empty for only \num{0.2}\% of indication edges. The section also rarely names even the family of a contraindicated disease: only \num{0.99}\% of contraindications are named at the family level there, against \num{11.23}\% of indications in the Indications section. The recall check of \cref{app:relcheck} confirms that the section does not state these facts in other words: annotators find \num{92}\% of the missed facts truly absent from it.

\paragraph{Where safety facts sit.}
In the relation check, most label sentences that name a contraindicated disease come from the Boxed Warning, Warnings and Precautions and Adverse Reactions sections. They read as warnings, such as an increased risk of a condition, rather than as formal contraindications (\cref{tab:audit}). This is why the whole label, not the Contraindications section, matches PrimeKG's contraindication relation.

\subsection{Chance Floor in the Label}

Labels are longer than Wikipedia articles and list many conditions in their adverse reaction tables. Their floor is therefore higher: an unrelated disease's exact name appears in the whole label for \num{2.10}\% of indication draws and \num{2.82}\% of contraindication draws. This floor is still an order of magnitude below the naming rate of real edges, and it is close for the two relations, so it cannot explain why the label shows no gap.

\subsection{Label Strata}

With the edge count held fixed, the whole label names contraindications slightly more often than indications (\cref{tab:logit}, coefficient \num{+0.304}, $p=\num{5.0e-3}$). In the strata, the gap is negative in all three (\cref{tab:strata}). The label is the only corpus in which a drug's safety facts are written down at least as often as its uses.

%% file: Appendix/H_probe.tex
\section{Closed-Book Probe}
\label{app:probe}

This appendix gives the format of the closed-book probe and its full results across five model sizes.

\subsection{Format}

\paragraph{Question format.}
We score each edge as a completion. The prompt is a short stem for the relation that names the drug and stops before the disease (\cref{tab:stems}). We append each option as a continuation and score it by the mean log-probability of its tokens. The option with the highest score is the model's answer. We use no chat template, passage or retrieval. The mean, rather than the sum, keeps long disease names from losing only because they have more tokens.

\begin{table}[h]
\centering
\footnotesize
\setlength{\tabcolsep}{3pt}
\begin{tabular}{>{\raggedright\arraybackslash}p{2.15cm}>{\raggedright\arraybackslash}p{4.95cm}}
\toprule
Relation & Prompt stem (X is the drug's name) \\
\midrule
Indication & \textit{X is a drug used to treat the disease called} \\
Contraindication & \textit{X is a drug that must not be given to patients who have} \\
Off-label use & \textit{X is a drug sometimes used off-label to treat} \\
Treats (Hetionet) & \textit{X is a drug used to treat the disease called} \\
Palliates (Hetionet) & \textit{X is a drug used to relieve the symptoms of} \\
\bottomrule
\end{tabular}
\caption{Prompt stems for the closed-book probe. We score each option as the continuation of the stem.}
\label{tab:stems}
\end{table}

{\scriptsize
\begin{verbatim}
def probe(model, stem, options):
    scores = []
    for o in options:
        ids = tok(stem + " " + o)
        n = len(tok(" " + o))
        lp = log_softmax(model(ids).logits)
        s = mean(lp[t-1, ids[t]]
                 for t in last(n, ids))
        scores.append(s)
    return options[argmax(scores)]
\end{verbatim}
}

\paragraph{Options.}
Each question has five options: the true disease and four wrong diseases drawn at random from the same relation's targets in the sample. The wrong options never include a disease that the drug is paired with in that relation within the sample. We cannot exclude true partners outside the sample. With about \num{600} targets per relation and a handful of partners per drug, the chance that a wrong option is an unrecorded true partner is about one percent, and it is the same for all relations. We shuffle the order of the options with a fixed seed. Chance accuracy is \num{20}\%.

\paragraph{Models.}
We probe Qwen2.5-0.5B, Qwen2.5-1.5B, Qwen2.5-3B, Qwen2.5-7B and Qwen2.5-14B, all instruction-tuned \citep{qwen2025qwen25}. All models run in \num{32}-bit floating point on CPU with Hugging Face Transformers. Every model answers the same \num{1800} questions with the same options in the same order, so we can pair them edge by edge.

\subsection{Results}

\paragraph{Accuracy at every size.}
\Cref{tab:probe} gives closed-book accuracy for every Qwen2.5 size, with the gap and its interval over drugs. Indication accuracy rises from \num{31.50}\% to about \num{50}\% and levels off from Qwen2.5-7B. Contraindication accuracy stays between \num{22.17}\% and \num{27.17}\%, close to chance at every size.

\begin{table}[h]
\centering
\footnotesize
\setlength{\tabcolsep}{2.1pt}
\begin{tabular}{lcccc}
\toprule
Model & Ind. & Off-label & Contra. & Gap [\num{95}\% CI] \\
\midrule
Qwen2.5-0.5B & \num{31.50} & \num{28.50} & \num{22.17} & \num{+9.3} [\num{4.0}, \num{14.9}] \\
Qwen2.5-1.5B & \num{45.67} & \num{35.83} & \num{25.33} & \num{+20.3} [\num{15.2}, \num{26.2}] \\
Qwen2.5-3B & \num{45.33} & \num{32.67} & \num{24.17} & \num{+21.2} [\num{16.1}, \num{26.9}] \\
Qwen2.5-7B & \num{50.83} & \num{35.00} & \num{23.83} & \num{+27.0} [\num{21.6}, \num{32.4}] \\
Qwen2.5-14B & \num{49.83} & \num{35.83} & \num{27.17} & \num{+22.7} [\num{17.1}, \num{28.3}] \\
\midrule
Wikipedia names it & \num{16.83} & \num{10.50} & \num{6.67} & \\
\bottomrule
\end{tabular}
\caption{Closed-book accuracy (\%) on the probe sample, \num{600} edges per relation, with chance at \num{20}\%. \emph{Gap} is indication minus contraindication accuracy, with a \num{95}\% bootstrap interval over drugs. The last row gives the share of the same edges that Wikipedia names (level \num{0} or \num{1}).}
\label{tab:probe}
\end{table}

\paragraph{Paired gain from scale.}
\Cref{tab:probe_full} gives the accuracy of the smallest and the 7B model with Wilson intervals, and the paired gain between them. Indications gain \num{19.33} points, off-label uses \num{6.50} points and contraindications only \num{1.67} points. The difference between the indication and contraindication gains is \num{17.67} points, with a bootstrap interval over \num{709} drugs that excludes zero by a wide margin.

\begin{table}[h]
\centering
\footnotesize
\setlength{\tabcolsep}{2pt}
\begin{tabular}{lccc}
\toprule
Relation & 0.5B [\num{95}\% CI] & 7B [\num{95}\% CI] & Gain \\
\midrule
Indication & \num{31.50} [\num{27.91}, \num{35.32}] & \num{50.83} [\num{46.84}, \num{54.82}] & \num{+19.33} \\
Off-label & \num{28.50} [\num{25.03}, \num{32.24}] & \num{35.00} [\num{31.29}, \num{38.90}] & \num{+6.50} \\
Contraind. & \num{22.17} [\num{19.03}, \num{25.66}] & \num{23.83} [\num{20.60}, \num{27.40}] & \num{+1.67} \\
\bottomrule
\end{tabular}
\caption{Probe accuracy (\%) of Qwen2.5-0.5B and Qwen2.5-7B, and the paired gain between them (points), on \num{600} edges per relation. Indication gain minus contraindication gain: \num{+17.67} [\num{12.68}, \num{22.70}], $p=\num{4.8e-12}$.}
\label{tab:probe_full}
\end{table}

\subsection{Accuracy by Whether Text Names the Fact}

\Cref{tab:probe_named} splits accuracy by whether Wikipedia names the fact. Scale helps most where text names the fact. For named indications, accuracy climbs from \num{57.4}\% to \num{84.2}\%. For named contraindications, it climbs from \num{22.5}\% to \num{40.0}\%, a smaller gain from a lower start. For facts that Wikipedia does not name, accuracy rises much less in every relation, and silent contraindications stay at chance from the smallest model to the largest.

\begin{table}[h]
\centering
\footnotesize
\setlength{\tabcolsep}{2.5pt}
\begin{tabular}{lcccccc}
\toprule
 & \multicolumn{2}{c}{Indication} & \multicolumn{2}{c}{Off-label} & \multicolumn{2}{c}{Contraind.} \\
\cmidrule(lr){2-3}\cmidrule(lr){4-5}\cmidrule(lr){6-7}
Model & named & not & named & not & named & not \\
\midrule
Qwen2.5-0.5B & \num{57.4} & \num{26.3} & \num{41.3} & \num{27.0} & \num{22.5} & \num{22.1} \\
Qwen2.5-1.5B & \num{74.3} & \num{39.9} & \num{61.9} & \num{32.8} & \num{27.5} & \num{25.2} \\
Qwen2.5-3B & \num{80.2} & \num{38.3} & \num{49.2} & \num{30.7} & \num{30.0} & \num{23.8} \\
Qwen2.5-7B & \num{84.2} & \num{44.1} & \num{57.1} & \num{32.4} & \num{32.5} & \num{23.2} \\
Qwen2.5-14B & \num{84.2} & \num{42.9} & \num{63.5} & \num{32.6} & \num{40.0} & \num{26.2} \\
\midrule
$n$ & \num{101} & \num{499} & \num{63} & \num{537} & \num{40} & \num{560} \\
\bottomrule
\end{tabular}
\caption{Closed-book accuracy (\%) split by whether Wikipedia names the fact (level \num{0} or \num{1}). The last row gives the number of edges in each group.}
\label{tab:probe_named}
\end{table}

From Qwen2.5-0.5B to Qwen2.5-14B, named contraindications gain \num{17.5} points [\num{4.9}, \num{30.8}], while unnamed contraindications gain only \num{4.1} points [\num{0.9}, \num{7.2}]. Even among named facts, Qwen2.5-14B gets \num{84.2}\% of indications but only \num{40.0}\% of contraindications. A named contraindication is usually named once, in a warning, while a named indication is usually the first thing an article says about the drug.

\paragraph{Why so few named contraindications.}
Only \num{40} of \num{600} sampled contraindications are named in Wikipedia, against \num{101} indications. Scale can only lift what text provides, and text provides few contraindications. This is why the contraindication curve in \cref{fig:scale} stays flat.

%% file: Appendix/I_page.tex
\section{Harder Wrong Options and the Drug's Page}
\label{app:page}

This appendix describes the harder-option test and the page test, and gives every per-model result behind \cref{fig:failure} and \cref{fig:retrieval}.

\subsection{Format}

\paragraph{Questions.}
This test reuses the question set of our companion study on declining. An indication question asks \emph{``Which disease is X used to treat?''}, and a contraindication question asks \emph{``In patients with which disease must X not be given?''}. Each question lists five diseases and the option \emph{Not enough information to tell}, lettered A to F. The position of every option, including the decline option, is random. The prompt has the shape below. We show one real question, with its options in an illustrative order and two wrong options left out.

\begin{quote}
\footnotesize\ttfamily
[Passage: <the drug's page, if any>]\\[2pt]
In patients with which disease must Ciprofloxacin not be given?\\
A. myasthenia gravis with thymus hyperplasia\\
B. <wrong option>\\
C. adult-onset myasthenia gravis\\
D. Not enough information to tell\\
E. myasthenia, limb-girdle, autoimmune\\
F. <wrong option>\\
Answer with the letter of one option only.
\end{quote}

Here the true answer, \emph{adult-onset myasthenia gravis}, sits among other forms of myasthenia. A model that knows only that ciprofloxacin is risky in myasthenia gravis cannot tell these options apart, which is the point of the test.

\paragraph{Wrong options.}
We take the wrong options first from the true disease's MONDO siblings, then from its cousins (other children of its grandparent), and only then from random diseases of the same relation. The wrong options never include the true disease's own ancestors or descendants, or any other disease the drug has under the same relation, since those options would also be correct. In \num{481} of the \num{856} questions, all four wrong options are siblings or cousins.

{\scriptsize
\begin{verbatim}
def wrong_options(d, r, y, k=4):
    bad = ({y} | ancestors(y) | descendants(y)
           | partners(d, r))
    out = []
    for pool in (siblings(y), cousins(y),
                 targets(r)):
        for z in shuffle(pool, seed=0):
            if z not in bad and z not in out:
                out.append(z)
            if len(out) == k:
                return out
\end{verbatim}
}

\paragraph{How often all options are siblings.}
\Cref{tab:sibshare} gives the share of questions whose four wrong options are all siblings. The share is highest in the family-only group, because those diseases sit in dense parts of MONDO. It is similar for the two relations within each group, so the option pool does not favor one relation.

\begin{table}[h]
\centering
\footnotesize
\begin{tabular}{lcc}
\toprule
Naming group & Indication & Contraindication \\
\midrule
Names the disease & \num{39.7} & \num{29.5} \\
Names only its family & \num{83.5} & \num{84.3} \\
Silent & \num{60.7} & \num{44.8} \\
\bottomrule
\end{tabular}
\caption{Share of questions (\%) whose four wrong options are all MONDO siblings of the answer.}
\label{tab:sibshare}
\end{table}

\paragraph{The page.}
A full drug article is too long for a prompt, so we cut a window around the evidence. The page starts with the first \num{800} characters of the drug's cleaned text, which is its Wikipedia article followed by its PubMed abstracts. We then add a window of \num{280} characters on each side of every mention of the answer or its family, merge windows that overlap, and stop at \num{1800} characters. A silent page has no mentions, so it is just the opening of the text. Every model sees the same page for the same question.

{\scriptsize
\begin{verbatim}
def page(text, names, head=800, w=280,
         cap=1800):
    spans = [(0, head)]
    for m in find_all(text, names):
        spans.append((m.start - w, m.end + w))
    spans = merge_overlaps(sorted(spans))
    out = " ... ".join(text[a:b]
                       for a, b in spans)
    return out[:cap]
\end{verbatim}
}

The page does not use the wrong options. It mentions only the answer and its family, as a retriever centered on the drug would. A page could still name a sibling by chance, and this is part of what makes family-only pages hard.

\paragraph{Scoring.}
We format each prompt with the model's chat template and read the answer from the next-token probability of each option letter. We score both spellings of each letter, with and without a leading space, keep the better one, and renormalize over the six letters. We batch prompts with left padding and rebuild the position indices from the attention mask, so batched and single scores agree.

{\scriptsize
\begin{verbatim}
def letter_answer(model, prompt):
    ids = chat_template(prompt)
    lp = log_softmax(model(ids).logits[-1])
    s = {L: max(lp[tok(L)], lp[tok(" " + L)])
         for L in "ABCDEF"}
    p = softmax(list(s.values()))
    return argmax(p), p
\end{verbatim}
}

\paragraph{Models.}
We test Qwen2.5-0.5B and Qwen2.5-7B \citep{qwen2025qwen25}, OLMo-2-7B \citep{olmo2024olmo2} and Granite-3.1-8B \citep{granite2024}, all instruction-tuned and in \num{32}-bit floating point on CPU. The three 7-8B models come from three different families with different training data.

\subsection{Closed Book}

\paragraph{The gap with harder options.}
\Cref{tab:distractor_full} repeats \cref{tab:distractors} and adds the gap computed within naming groups and its $p$-value. The within-group gap removes the effect of the two relations having different mixes of named and silent questions. Both tests agree in every model.

\begin{table}[h]
\centering
\footnotesize
\setlength{\tabcolsep}{3pt}
\begin{tabular}{lccccc}
\toprule
Model & Ind. & Contra. & Gap & Within & $p$ (within) \\
\midrule
Qwen2.5-0.5B & \num{36.0} & \num{28.1} & \num{+7.9} & \num{+8.2} & \num{8.9e-3} \\
Qwen2.5-7B & \num{56.0} & \num{25.1} & \num{+30.9} & \num{+31.3} & $<\num{e-15}$ \\
OLMo-2-7B & \num{56.0} & \num{33.0} & \num{+23.0} & \num{+23.6} & \num{2.3e-13} \\
Granite-3.1-8B & \num{58.7} & \num{12.8} & \num{+45.9} & \num{+46.3} & $<\num{e-15}$ \\
\bottomrule
\end{tabular}
\caption{Harder-option test, closed book (\%); \num{450} indication and \num{406} contraindication questions. The $p$-values of the raw gap are \num{0.013} for Qwen2.5-0.5B, \num{1.5e-11} for OLMo-2-7B, and below \num{e-15} for Qwen2.5-7B and Granite-3.1-8B.}
\label{tab:distractor_full}
\end{table}

\paragraph{How models fail.}
\Cref{tab:failure} gives the share of wrong answers and declines, and the precision on answered questions, for each model. Two kinds of failure appear. Qwen2.5-0.5B and OLMo-2-7B rarely decline and answer most contraindication questions wrongly. Qwen2.5-7B and Granite-3.1-8B decline on half or more of contraindication questions. Their precision on answered contraindications, \num{51.0}\% and \num{39.4}\%, is still far below their precision on indications.

\begin{table}[h]
\centering
\footnotesize
\setlength{\tabcolsep}{2.6pt}
\begin{tabular}{lcccccc}
\toprule
 & \multicolumn{3}{c}{Indication} & \multicolumn{3}{c}{Contraindication} \\
\cmidrule(lr){2-4}\cmidrule(lr){5-7}
Model & wrong & decl. & prec. & wrong & decl. & prec. \\
\midrule
Qwen2.5-0.5B & \num{60.7} & \num{3.3} & \num{37.2} & \num{71.4} & \num{0.5} & \num{28.2} \\
Qwen2.5-7B & \num{22.0} & \num{22.0} & \num{71.8} & \num{24.1} & \num{50.7} & \num{51.0} \\
OLMo-2-7B & \num{34.0} & \num{10.0} & \num{62.2} & \num{64.0} & \num{3.0} & \num{34.0} \\
Granite-3.1-8B & \num{28.7} & \num{12.7} & \num{67.2} & \num{19.7} & \num{67.5} & \num{39.4} \\
\bottomrule
\end{tabular}
\caption{How models fail closed book with harder wrong options (\%). \emph{Decl.} is the share of questions answered \emph{Not enough information}; \emph{prec.} is the accuracy on the questions the model answers.}
\label{tab:failure}
\end{table}

\paragraph{Closed book by naming group.}
The \emph{Closed} column of \cref{tab:page_full} shows that closed-book accuracy also follows naming. For Qwen2.5-7B, named indications reach \num{78.7}\% and silent ones \num{39.3}\%. Named contraindications reach only \num{33.3}\% and silent ones \num{17.3}\%. The model does best on the facts that text names, and text names few contraindications.

\subsection{With the Page}

\Cref{tab:page_full} gives accuracy closed book and with the page for every model and naming group, with bootstrap intervals over questions. \Cref{tab:retrieval} summarizes the changes. When the page names the answer, it helps both relations in every model, and it helps contraindications at least as much as indications. When the page names only the family or nothing, it helps little or hurts, because the larger models decline more.

\begin{table*}[t]
\centering
\footnotesize
\setlength{\tabcolsep}{2.6pt}
\begin{tabular}{llrccc@{\hskip 8pt}llrccc}
\toprule
\multicolumn{6}{c}{Indication} & \multicolumn{6}{c}{Contraindication} \\
\cmidrule(lr){1-6}\cmidrule(lr){7-12}
Model & Group & $n$ & Closed & Page & Change [\num{95}\% CI] & Model & Group & $n$ & Closed & Page & Change [\num{95}\% CI] \\
\midrule
Qwen2.5-0.5B & named & \num{150} & \num{47.3} & \num{86.7} & \num{+39.3} [\num{+31.3}, \num{+47.3}] & Qwen2.5-0.5B & named & \num{150} & \num{30.0} & \num{70.7} & \num{+40.7} [\num{+32.0}, \num{+49.3}] \\
 & family & \num{150} & \num{34.0} & \num{44.7} & \num{+10.7} [\num{+2.0}, \num{+20.0}] & & family & \num{106} & \num{23.6} & \num{27.4} & \num{+3.8} [\num{-4.7}, \num{+12.3}] \\
 & silent & \num{150} & \num{26.7} & \num{34.0} & \num{+7.3} [\num{-0.7}, \num{+15.3}] & & silent & \num{150} & \num{29.3} & \num{23.3} & \num{-6.0} [\num{-13.3}, \num{+0.7}] \\
\midrule
Qwen2.5-7B & named & \num{150} & \num{78.7} & \num{85.3} & \num{+6.7} [\num{0.0}, \num{+13.3}] & Qwen2.5-7B & named & \num{150} & \num{33.3} & \num{44.0} & \num{+10.7} [\num{+0.7}, \num{+20.0}] \\
 & family & \num{150} & \num{50.0} & \num{38.0} & \num{-12.0} [\num{-18.7}, \num{-5.3}] & & family & \num{106} & \num{24.5} & \num{17.0} & \num{-7.5} [\num{-17.9}, \num{+2.8}] \\
 & silent & \num{150} & \num{39.3} & \num{22.7} & \num{-16.7} [\num{-24.0}, \num{-10.0}] & & silent & \num{150} & \num{17.3} & \num{3.3} & \num{-14.0} [\num{-20.0}, \num{-8.0}] \\
\midrule
OLMo-2-7B & named & \num{150} & \num{75.3} & \num{88.0} & \num{+12.7} [\num{+5.3}, \num{+20.0}] & OLMo-2-7B & named & \num{150} & \num{42.0} & \num{54.7} & \num{+12.7} [\num{+3.3}, \num{+22.0}] \\
 & family & \num{150} & \num{48.0} & \num{45.3} & \num{-2.7} [\num{-10.7}, \num{+5.3}] & & family & \num{106} & \num{30.2} & \num{30.2} & \num{0.0} [\num{-8.5}, \num{+8.5}] \\
 & silent & \num{150} & \num{44.7} & \num{43.3} & \num{-1.3} [\num{-9.3}, \num{+6.7}] & & silent & \num{150} & \num{26.0} & \num{20.7} & \num{-5.3} [\num{-12.7}, \num{+1.3}] \\
\midrule
Granite-3.1-8B & named & \num{150} & \num{76.7} & \num{90.7} & \num{+14.0} [\num{+7.3}, \num{+20.7}] & Granite-3.1-8B & named & \num{150} & \num{18.7} & \num{38.0} & \num{+19.3} [\num{+10.7}, \num{+28.0}] \\
 & family & \num{150} & \num{54.0} & \num{44.0} & \num{-10.0} [\num{-17.3}, \num{-2.7}] & & family & \num{106} & \num{9.4} & \num{17.9} & \num{+8.5} [\num{0.0}, \num{+17.0}] \\
 & silent & \num{150} & \num{45.3} & \num{29.3} & \num{-16.0} [\num{-24.0}, \num{-8.0}] & & silent & \num{150} & \num{9.3} & \num{2.0} & \num{-7.3} [\num{-12.0}, \num{-2.7}] \\
\midrule
Pooled 7-8B & named & \num{150} & & & \num{+11.1} [\num{+6.7}, \num{+15.8}] & Pooled 7-8B & named & \num{150} & & & \num{+14.2} [\num{+7.3}, \num{+20.9}] \\
 & family & \num{150} & & & \num{-8.2} [\num{-13.3}, \num{-3.1}] & & family & \num{106} & & & \num{+0.3} [\num{-6.3}, \num{+6.9}] \\
 & silent & \num{150} & & & \num{-11.3} [\num{-17.1}, \num{-6.0}] & & silent & \num{150} & & & \num{-8.9} [\num{-12.9}, \num{-5.1}] \\
\bottomrule
\end{tabular}
\caption{Accuracy (\%) closed book and with the drug's page, and the change (points) with a \num{95}\% bootstrap interval over questions, by what the page says about the answer. The pooled rows average each question over Qwen2.5-7B, OLMo-2-7B and Granite-3.1-8B first. In the named group, the indication change minus the contraindication change is \num{-3.1} [\num{-11.6}, \num{+5.1}] points pooled, so a written safety fact helps at least as much as a written benefit fact.}
\label{tab:page_full}
\end{table*}

\begin{table}[h]
\centering
\footnotesize
\setlength{\tabcolsep}{2.6pt}
\begin{tabular}{lcccccc}
\toprule
 & \multicolumn{2}{c}{Names it} & \multicolumn{2}{c}{Family only} & \multicolumn{2}{c}{Silent} \\
\cmidrule(lr){2-3}\cmidrule(lr){4-5}\cmidrule(lr){6-7}
Model & Ind. & Con. & Ind. & Con. & Ind. & Con. \\
\midrule
Qwen2.5-0.5B & \num{+39.3} & \num{+40.7} & \num{+10.7} & \num{+3.8} & \num{+7.3} & \num{-6.0} \\
Qwen2.5-7B & \num{+6.7} & \num{+10.7} & \num{-12.0} & \num{-7.5} & \num{-16.7} & \num{-14.0} \\
OLMo-2-7B & \num{+12.7} & \num{+12.7} & \num{-2.7} & \num{0.0} & \num{-1.3} & \num{-5.3} \\
Granite-3.1-8B & \num{+14.0} & \num{+19.3} & \num{-10.0} & \num{+8.5} & \num{-16.0} & \num{-7.3} \\
\midrule
Pooled 7-8B & \num{+11.1} & \num{+14.2} & \num{-8.2} & \num{+0.3} & \num{-11.3} & \num{-8.9} \\
\bottomrule
\end{tabular}
\caption{Change in accuracy (points) from closed book to the drug's page, by what the page says about the answer. Each cell has \num{150} questions, except \num{106} family-only contraindication questions.}
\label{tab:retrieval}
\end{table}

\paragraph{The smallest model.}
Qwen2.5-0.5B gains the most from a naming page, about \num{40} points for both relations. It rarely declines, so when the page contains the answer it copies it. It is also the only model that gains on indications from family-only and silent pages, by \num{10.7} and \num{7.3} points.

\paragraph{Declines with the page.}
\Cref{tab:declines} gives the share of questions that the three 7-8B models answer with \emph{Not enough information}. A silent page raises declines the most, by \num{26.0} points on indications and \num{32.9} points on contraindications. A page that names the answer barely changes them. The models decline when the page gives them nothing and answer when it names the disease. With a silent page, Qwen2.5-7B declines on \num{94.0}\% of contraindication questions.

\begin{table}[h]
\centering
\footnotesize
\setlength{\tabcolsep}{3.5pt}
\begin{tabular}{lcccc}
\toprule
 & \multicolumn{2}{c}{Indication} & \multicolumn{2}{c}{Contraindication} \\
\cmidrule(lr){2-3}\cmidrule(lr){4-5}
Page & closed & page & closed & page \\
\midrule
Names the disease & \num{6.0} & \num{5.8} & \num{36.2} & \num{41.6} \\
Names only its family & \num{9.1} & \num{26.0} & \num{42.1} & \num{54.4} \\
Silent & \num{29.6} & \num{55.6} & \num{43.3} & \num{76.2} \\
\bottomrule
\end{tabular}
\caption{Share of questions answered \emph{Not enough information} (\%), closed book and with the drug's page, averaged over Qwen2.5-7B, OLMo-2-7B and Granite-3.1-8B.}
\label{tab:declines}
\end{table}

%% file: Appendix/J_relcheck.tex
\section{Does the Sentence State the Relation?}
\label{app:relcheck}

The main measurement counts a fact as named when the drug's text contains the disease's name. This appendix checks whether the sentence that contains the name also states the relation, with a model judge, a keyword check and four annotators.

\paragraph{Sentences.}
For every indication and contraindication edge and each corpus, we split the drug's text into sentences and keep up to three that contain the disease's exact or alternative name. We strip Wikipedia markup first. This sentence-level call agrees with the main measurement on \num{97.9}\% of Wikipedia edges, \num{99.3}\% of PubMed edges and all label edges.

{\scriptsize
\begin{verbatim}
def naming_sentences(text, y, k=3):
    text = strip_markup(text)
    out = []
    for s in split_sentences(text):
        if matches(s, [label(y)] + redirects(y)):
            out.append(s)
        if len(out) == k:
            break
    return out

def supports(rel, sentence, d, y):
    L = judge(sentence, d, y)  # A, B, C or D
    if rel == "indication":
        return L == "A"
    return L in ("B", "C")     # strict: "B"
\end{verbatim}
}

\paragraph{Model judge.}
Qwen2.5-7B reads the first naming sentence and picks one of four letters:
\begin{quote}
\footnotesize\ttfamily
Sentence: <sentence>\\[2pt]
What does this sentence say about X and Y?\\
A. X is used to treat or manage Y.\\
B. X must not be used in patients with Y.\\
C. X may cause or worsen Y, or needs caution in patients with Y.\\
D. The sentence mentions both but states no link between them.\\
Answer with the letter of one option only.
\end{quote}
We read the answer from the letter probabilities, as in \cref{app:page}. A sentence supports an indication when the judge picks A. It supports a contraindication when the judge picks B or C, since a warning about harm is a safety statement. The strict reading (\emph{strict} in \cref{tab:relcheck_full}) accepts only B.

\paragraph{Keyword check.}
As a cheap baseline, a sentence supports an indication when it contains a treatment word (\emph{treat}, \emph{indicated}, \emph{therapy}, \emph{used for}, \emph{management of}, \emph{prevent} and similar). It supports a contraindication when it contains a contraindication word (\emph{contraindicated}, \emph{must not}, \emph{should not}, \emph{avoid}, \emph{not recommended}) or a warning word (\emph{caution}, \emph{warning}, \emph{worsen}, \emph{exacerbate}, \emph{risk}, \emph{adverse}, \emph{may cause}).

\begin{table}[h]
\centering
\footnotesize
\setlength{\tabcolsep}{1.5pt}
\begin{tabular}{llccc}
\toprule
Corpus & Definition & Ind. & Contra. & OR [\num{95}\% CI] \\
\midrule
Wikipedia & named & \num{18.41} & \num{8.01} & \num{2.68} [\num{2.16}, \num{3.34}] \\
 & keyword & \num{16.32} & \num{4.18} & \num{4.60} [\num{3.49}, \num{6.07}] \\
 & judge & \num{17.19} & \num{7.38} & \num{2.70} [\num{2.15}, \num{3.39}] \\
 & strict & \num{17.19} & \num{1.10} & \num{18.68} [\num{11.43}, \num{30.54}] \\
\midrule
PubMed & named & \num{25.61} & \num{10.80} & \num{2.83} [\num{2.34}, \num{3.41}] \\
 & keyword & \num{13.59} & \num{2.73} & \num{5.64} [\num{4.08}, \num{7.78}] \\
 & judge & \num{17.83} & \num{9.35} & \num{2.09} [\num{1.71}, \num{2.57}] \\
 & strict & \num{17.83} & \num{0.06} & \num{460.00} [\num{58.01}, \num{3647.67}] \\
\midrule
Label & named & \num{23.73} & \num{22.88} & \num{1.05} [\num{0.88}, \num{1.25}] \\
 & keyword & \num{20.55} & \num{16.81} & \num{1.30} [\num{1.07}, \num{1.57}] \\
 & judge & \num{18.86} & \num{20.76} & \num{0.88} [\num{0.73}, \num{1.07}] \\
 & strict & \num{18.86} & \num{2.19} & \num{10.32} [\num{6.98}, \num{15.25}] \\
\bottomrule
\end{tabular}
\caption{Share of facts (\%) counted as stated under each definition, and the within-drug odds ratio with Robins-Breslow-Greenland intervals. \num{1722} edges per relation for Wikipedia and PubMed, \num{1416} for the label.}
\label{tab:relcheck_full}
\end{table}

\paragraph{Results.}
\Cref{tab:relcheck_full} gives every definition. With the judge, the prose gap stays at \num{2.09}-\num{2.70}, and the label stays even. The strict reading gives very large odds ratios in every corpus, the label included, because formal contraindication wording is rare everywhere: even the label states it for only \num{2.19}\% of contraindications. Most safety facts appear as warnings, so we use the judge with warnings as the main definition.

\paragraph{Annotation.}
We draw \num{160} named sentences with seed \num{0}: \num{40} for each relation from Wikipedia and \num{40} from the whole label. We shuffle them and hide the graph relation. Four annotators label each sentence on their own with one letter: T (the drug treats the disease), C (it must not be used), W (it may cause or worsen the disease, or needs caution) or M (a mention with no link). We take the majority label and leave out the \num{6} ties.

\begin{table}[h]
\centering
\footnotesize
\setlength{\tabcolsep}{2.5pt}
\begin{tabular}{llcccccc}
\toprule
Relation & Corpus & $n$ & T & C & W & M & Supports \\
\midrule
Indication & Wikipedia & \num{39} & \num{97.4} & \num{0.0} & \num{0.0} & \num{2.6} & \num{97.4} \\
 & Label & \num{40} & \num{80.0} & \num{0.0} & \num{10.0} & \num{10.0} & \num{80.0} \\
Contra. & Wikipedia & \num{39} & \num{5.1} & \num{10.3} & \num{82.1} & \num{2.6} & \num{92.3} \\
 & Label & \num{36} & \num{2.8} & \num{11.1} & \num{83.3} & \num{2.8} & \num{94.4} \\
\bottomrule
\end{tabular}
\caption{Majority label of four annotators (\%) on named sentences. \emph{Supports} is T for indications and C or W for contraindications.}
\label{tab:audit}
\end{table}

\paragraph{Agreement.}
The annotators agree closely: Fleiss' $\kappa=\num{0.89}$, pairwise Cohen's $\kappa$ from \num{0.86} to \num{0.93}, and all four agree on \num{141} of \num{160} sentences. Against the majority label, the judge agrees on \num{95.5}\% of sentences ($\kappa=\num{0.92}$) and on \num{98.1}\% of the supported or not supported calls ($\kappa=\num{0.88}$). The keyword check agrees on \num{64.3}\% ($\kappa=\num{0.43}$). It misses most safety statements because lists of adverse reactions, such as \emph{``Hematological: agranulocytosis, thrombocytopenia''}, contain no relation word. By the majority label, \num{88.6}\% of named indications and \num{93.3}\% of named contraindications state the graph's relation (\cref{tab:audit}).

\paragraph{What PrimeKG calls a contraindication.}
Only \num{10}-\num{11}\% of the contraindication sentences are formal contraindications (C). Most are warnings (W): an increased risk of myocardial infarction with ketorolac, or agranulocytosis in a list of adverse reactions. PrimeKG's contraindication relation therefore covers formal contraindications and safety warnings, and our results apply to both.

\paragraph{Recall check.}
To test whether the matcher misses the wording of the label, we draw \num{50} of the \num{1227} contraindications whose label has a non-empty Contraindications section that the matcher finds nothing in. The same four annotators read the section and mark each fact as Y (stated in other words), B (only a broader group is stated) or N (not there). They agree closely (Fleiss' $\kappa=\num{0.92}$, unanimous on \num{49} items). By the majority label, \num{45} facts (\num{91.8}\%) are not in the section, \num{3} (\num{6.1}\%) are stated in other words, and \num{1} is stated only as a broader group; \num{1} is a tie. The three misses are \emph{ulcerative colitis} for clindamycin, \emph{cardiac valvular disorders} for cabergoline, and \emph{marked hepatic damage} for sulfamethoxazole. The low coverage of the Contraindications section is a property of the section, not of the matcher.

\subsection{Annotation Examples}

\Cref{tab:audit_ex} shows sentences on which all four annotators agree, one for each label. The Diclofenac sentence comes from the label of a drug that treats rheumatoid arthritis, but the sentence itself only compares two groups of patients, so it does not state the relation. Such sentences are why we check the relation and not only the name.

\begin{table}[h]
\centering
\footnotesize
\setlength{\tabcolsep}{2.5pt}
\begin{tabular}{>{\raggedright\arraybackslash}p{2.05cm}>{\raggedright\arraybackslash}p{5.0cm}c}
\toprule
Drug, disease & Sentence & Label \\
\midrule
Isocarboxazid, endogenous depression & ``It has demonstrated efficacy in various depressive subtypes, including atypical depression, melancholic depression, and endogenous depression.'' & T \\
Isocarboxazid, hypertension & ``Marplan (isocarboxazid) should not be administered to any patient with a confirmed or suspected cerebrovascular defect or to any patient with cardiovascular disease, hypertension, \ldots'' & C \\
Ketorolac, myocardial infarction & ``Nonsteroidal anti-inflammatory drugs (NSAIDs) cause an increased risk of serious cardiovascular thrombotic events, including myocardial infarction and stroke, \ldots'' & W \\
Diclofenac, rheumatoid arthritis & ``Patients with osteoarthritis more often develop symptomatic liver disease than patients with rheumatoid arthritis.'' & M \\
\bottomrule
\end{tabular}
\caption{Sentences on which all four annotators agree. T: treats; C: must not be used; W: may cause or worsen, or needs caution; M: mention with no link.}
\label{tab:audit_ex}
\end{table}

\paragraph{Where annotators disagree.}
The \num{19} sentences without full agreement fall into three kinds. The first is a bare heading or list item with no verb, such as \emph{``Osteoporosis-related fractures''} for omeprazole or \emph{``Kidney or liver problems''} for rabeprazole. Most annotators read these as warnings, because they appear in warning sections, but one reads them as mentions. The second is the line between C and W, such as \emph{``Avoid use of fenoprofen calcium in patients with advanced renal disease unless benefits are expected to outweigh risk''}. Both labels support a contraindication, so this disagreement does not change any result. The third is a name used in another sense. For histrelin, \emph{hypertension} appears inside \emph{idiopathic intracranial hypertension}, a different disease. For norepinephrine, the sentence says that alpha blockers treat hypertension, not that norepinephrine does. The relation check exists to catch such cases.

\paragraph{Judge errors.}
The judge disagrees with the majority on \num{4.5}\% of sentences. Two cases show what can go wrong. For doripenem, a label fragment lists pyelonephritis among approved uses, and for ipratropium, a sentence states its use in the common cold, yet the judge answers \emph{none} for both. These errors drop real indications, so they shrink the gap rather than create it.

%% file: Appendix/K_routing.tex
\section{Routing Test}
\label{app:routing}

\paragraph{Setup.}
We ask the harder-option questions of \cref{app:page} a third time. The questions and options are built exactly as before, so they are the same questions. Only the passage changes: we cut it from the drug's whole FDA label, with the same window rule of \num{800} opening characters and windows around each mention, up to \num{1800} characters. We leave out the questions whose drug has no label after selection, which keeps the other questions identical. This leaves \num{726} questions: \num{380} indications and \num{346} contraindications. The \emph{routed} condition takes the prose answer for indications and the label answer for contraindications. In a deployed agent, the planner would classify the relation from the question. Here the relation is known, so the test measures the value of the route, not the accuracy of the planner.

{\scriptsize
\begin{verbatim}
def routed_answer(model, q):
    if q.relation == "indication":
        page = page(prose(q.drug), names(q.y))
    else:
        page = page(label(q.drug), names(q.y))
    return letter_answer(model,
                         page + q.text)
\end{verbatim}
}

\begin{table}[h]
\centering
\footnotesize
\setlength{\tabcolsep}{3pt}
\begin{tabular}{llccc}
\toprule
Model & Condition & Correct & Wrong & Decline \\
\midrule
Qwen2.5-7B & prose & \num{37.5} & \num{8.3} & \num{54.3} \\
 & label & \num{32.0} & \num{5.1} & \num{62.9} \\
 & routed & \num{38.2} & \num{7.9} & \num{54.0} \\
OLMo-2-7B & prose & \num{49.3} & \num{29.5} & \num{21.2} \\
 & label & \num{50.3} & \num{28.0} & \num{21.8} \\
 & routed & \num{50.6} & \num{29.5} & \num{20.0} \\
Granite-3.1-8B & prose & \num{39.9} & \num{12.8} & \num{47.2} \\
 & label & \num{38.7} & \num{9.4} & \num{51.9} \\
 & routed & \num{42.6} & \num{12.9} & \num{44.5} \\
\midrule
Pooled & closed book & \num{43.1} & \num{30.8} & \num{26.1} \\
 & prose & \num{42.2} & \num{16.9} & \num{40.9} \\
 & label & \num{40.3} & \num{14.1} & \num{45.5} \\
 & routed & \num{43.8} & \num{16.8} & \num{39.5} \\
\bottomrule
\end{tabular}
\caption{Outcomes (\%) on all \num{726} questions with a label page.}
\label{tab:route_full}
\end{table}

\paragraph{Results.}
\Cref{tab:route_full} gives the outcomes for each model. Routing raises correct answers in every model, by \num{0.7} to \num{2.7} points, and by \num{1.5} [\num{-0.2}, \num{+3.2}] points pooled. On contraindications alone, the label page raises correct answers by \num{3.2} [\num{-0.4}, \num{+6.6}] points and lowers declines by \num{3.0} points. On indications, it lowers correct answers by \num{6.6} [\num{-9.7}, \num{-3.4}] points and raises declines by \num{11.6} points, so routing indications to the label would hurt. The split in \cref{tab:route} and \cref{tab:route_split} explains the small overall gain. The label page helps safety questions by \num{15.8} points when the whole label names the answer, and it cannot help the \num{61}\% of contraindication questions whose label does not.

\begin{table}[h]
\centering
\footnotesize
\setlength{\tabcolsep}{1.8pt}
\begin{tabular}{llrccc}
\toprule
Relation & Named & $n$ & Correct & Wrong & Decline \\
\midrule
Ind. & yes & \num{117} & \num{+3.4} [\num{-1.7}, \num{8.5}] & \num{-1.7} & \num{-1.7} \\
 & no & \num{263} & \num{-11.0} [\num{-15.1}, \num{-7.1}] & \num{-6.5} & \num{+17.5} \\
Contra. & yes & \num{135} & \num{+15.8} [\num{8.6}, \num{23.2}] & \num{-3.7} & \num{-12.1} \\
 & no & \num{211} & \num{-4.9} [\num{-8.2}, \num{-1.7}] & \num{+2.1} & \num{+2.8} \\
\bottomrule
\end{tabular}
\caption{Label page minus prose page (points), pooled over the three 7-8B models, split by whether the whole label names the answer (\emph{Named}). Brackets give \num{95}\% bootstrap intervals over questions.}
\label{tab:route_split}
\end{table}

\paragraph{When the label names the answer.}
\Cref{tab:route_split} gives the full split. When the whole label names a contraindication, the label page raises correct answers from \num{40.0}\% to \num{55.8}\% and cuts declines from \num{45.2}\% to \num{33.1}\%. When it does not, the label page helps nothing, and correct answers fall slightly. For indications, the label rarely adds anything that prose lacks: when it names the answer, prose usually names it too, and prose already reaches \num{84.3}\%. When it does not, the label page is mostly dosing and warnings, and the models decline more.

\paragraph{What this means for the router.}
The router should send safety questions to the label, and benefit questions to prose. It should also check whether the label names the answer before it lets the model answer. On the \num{61}\% of contraindication questions whose label is silent, no page we test helps, and the right action is to decline and name the silent source. \Cref{app:cases} shows questions that routing fixes.

%% file: Appendix/L_hetionet.tex
\section{Negative Control on Hetionet}
\label{app:hetionet}

Hetionet's \emph{treats} and \emph{palliates} relations both describe benefit. If our matcher created a gap between any two relations, it would create one here too. We run the same measurement on the same corpus, using the Disease Ontology instead of MONDO for the disease hierarchy.

\paragraph{Setup.}
We draw \num{600} \emph{treats} edges and all \num{385} measurable \emph{palliates} edges (\cref{app:data}) and fetch the drugs' Wikipedia articles, \num{497} in all. The matcher, the names, the floor and the tests are the same as for PrimeKG. The only change is the hierarchy: the Disease Ontology gives \num{103} family links over \num{74} diseases.

\paragraph{Result.}
Wikipedia names the disease for \num{57.33}\% of \num{600} sampled \emph{treats} edges and \num{56.10}\% of \num{385} measurable \emph{palliates} edges, an odds ratio of \num{1.05} ($p=\num{0.70}$). Within the \num{51} drugs that have both relations, the Mantel-Haenszel odds ratio is \num{2.48} [\num{1.21}, \num{5.08}]. This within-drug difference does not survive the edge-count control. With the drug's edge count held fixed, the relation term is not significant ($p=\num{0.16}$), and the strata change sign (\cref{tab:hetionet}). In PrimeKG, the relation term stays significant under the same control ($p=\num{2.4e-4}$ in Wikipedia).

\begin{table}[h]
\centering
\footnotesize
\setlength{\tabcolsep}{3pt}
\begin{tabular}{lcc}
\toprule
 & \emph{treats} & \emph{palliates} \\
\midrule
Named in Wikipedia (\%) & \num{57.33} ($n=\num{600}$) & \num{56.10} ($n=\num{385}$) \\
Odds ratio, pooled & \multicolumn{2}{c}{\num{1.05}, $p=\num{0.70}$} \\
Within drug (\num{51} drugs) & \multicolumn{2}{c}{\num{2.48} [\num{1.21}, \num{5.08}]} \\
Relation term, edges fixed & \multicolumn{2}{c}{$p=\num{0.16}$} \\
\midrule
Gap, \num{1}-\num{2} edges & \multicolumn{2}{c}{\num{+8.16} ($p=\num{0.06}$)} \\
Gap, \num{3}-\num{5} edges & \multicolumn{2}{c}{\num{+0.93} ($p=\num{0.89}$)} \\
Gap, \num{6}-\num{12} edges & \multicolumn{2}{c}{\num{-11.38} ($p=\num{0.32}$)} \\
\bottomrule
\end{tabular}
\caption{Negative control on Hetionet: two benefit relations, the same corpus and the same matcher. No gap survives the edge-count control, and the strata change sign.}
\label{tab:hetionet}
\end{table}

\paragraph{Why the within-drug ratio is above one.}
The within-drug odds ratio rests on only the \num{51} drugs that have both relations, and its interval is wide. It does not survive the edge-count control ($p=\num{0.16}$). \Cref{tab:het_strata} shows why: the gap is \num{+8.16} points for drugs with one or two edges, close to zero for three to five, and reversed for six to twelve. A matcher bias would push the same way in every stratum. In PrimeKG, the same control leaves the gap in place in every prose corpus (\cref{tab:logit}).

\begin{table}[h]
\centering
\footnotesize
\setlength{\tabcolsep}{3pt}
\begin{tabular}{lcccc}
\toprule
Edges per drug & \emph{treats} & $n$ & \emph{palliates} & $n$ \\
\midrule
\num{1}-\num{2} & \num{68.06} & \num{310} & \num{59.91} & \num{217} \\
\num{3}-\num{5} & \num{51.61} & \num{93} & \num{50.68} & \num{146} \\
\num{6}-\num{12} & \num{43.17} & \num{139} & \num{54.55} & \num{22} \\
\bottomrule
\end{tabular}
\caption{Hetionet naming rate (\%) in Wikipedia within strata of the drug's edge count.}
\label{tab:het_strata}
\end{table}

\paragraph{Naming levels.}
Hetionet's facts are named far more often than PrimeKG's: \num{54.5}\% of its sampled edges are named exactly, against \num{13.2}\% in the PrimeKG text sample (\cref{tab:sampletiers}). Hetionet covers fewer and broader diseases, such as \emph{hypertension} and \emph{asthma}, which articles name directly. This does not affect the control, which compares two relations on the same diseases and the same articles.

%% file: Appendix/M_cases.tex
\section{Case Studies}
\label{app:cases}

This appendix follows single facts through the study, so the reader can see what each measurement does on real text. All examples come from our released files.

\subsection{One Drug through Every Stage}

Ciprofloxacin is the running example of the paper. The text sample holds three of its indications and three of its contraindications. \Cref{tab:cipro} shows how plainly each corpus names each fact.

\begin{table}[h]
\centering
\footnotesize
\setlength{\tabcolsep}{2.5pt}
\begin{tabular}{>{\raggedright\arraybackslash}p{2.6cm}ccc}
\toprule
Fact & Wikipedia & PubMed & Label \\
\midrule
\multicolumn{4}{l}{\emph{Indications}} \\
infectious diarrheal disease & none & none & none \\
shigellosis & none & named & none \\
conjunctivitis & none & none & named \\
\midrule
\multicolumn{4}{l}{\emph{Contraindications}} \\
peripheral neuropathy & none & named & named \\
adult-onset myasthenia gravis & family & family & family \\
periarthritis & family & family & family \\
\bottomrule
\end{tabular}
\caption{How plainly each corpus names six ciprofloxacin facts. \emph{Named} is level \num{0} or \num{1}; \emph{family} is level \num{2}.}
\label{tab:cipro}
\end{table}

\paragraph{Indications.}
No corpus names \emph{infectious diarrheal disease}, because text names the specific infections instead. PubMed names shigellosis in a sentence that ends \emph{``\ldots with endemic and epidemic shigellosis and invasive salmonellosis''}, and the judge reads it as a treatment with probability \num{0.995}. Only the label names conjunctivitis, in the Indications section of the ophthalmic solution.

\paragraph{Contraindications.}
PubMed names peripheral neuropathy in a warning, \emph{``\ldots rupture, peripheral neuropathy and aortic aneurysm''}, and the label names it in the Boxed Warning: \emph{``WARNING: SERIOUS ADVERSE REACTIONS INCLUDING TENDINITIS, TENDON RUPTURE, PERIPHERAL NEUROPATHY, CENTRAL NERVOUS SYSTEM EFFECTS AND EXACERBATION OF MYASTHENIA GRAVIS''}. The same warning names \emph{myasthenia gravis}, but PrimeKG records the fact as \emph{adult-onset myasthenia gravis}, a child of that term. Every corpus therefore names it only at the family level. Periarthritis is the same: text names only a broader term, not this specific disease.

\paragraph{What the models do.}
The harder-option question for adult-onset myasthenia gravis puts the answer next to other forms of myasthenia (\cref{app:page}). Closed book, Qwen2.5-7B picks \emph{myasthenia gravis with thymus hyperplasia}, and it keeps this answer with the prose page and with the label page. Closed book, Qwen2.5-0.5B picks \emph{myasthenia, limb-girdle, autoimmune}. Granite-3.1-8B is wrong closed book and right with the prose page. OLMo-2-7B is wrong with the prose page and right with the label page. For periarthritis, Qwen2.5-7B and Granite-3.1-8B decline, and OLMo-2-7B and Qwen2.5-0.5B pick \emph{reactive arthritis}. A family-level mention points the models to the right group but not to the right member, which is the behavior that the family-only group of \cref{tab:retrieval} measures.

\paragraph{The lesson.}
A clinician would say that the label covers the myasthenia gravis risk. Our measurement agrees that the label points there, and it records the fact as family only because the graph is more specific than the text. This is the source of the graph-specificity limitation in \cref{sec:limits}.

\subsection{Safety Facts Only the Label Names}

\Cref{tab:labelonly} lists contraindications that neither Wikipedia nor PubMed names but the whole label does. The judge reads the label sentences for the first three as warnings. These facts are among the \num{11.4}\% of prose misses that the label rescues (\cref{tab:rescue}).

\begin{table}[h]
\centering
\footnotesize
\setlength{\tabcolsep}{3pt}
\begin{tabular}{ll}
\toprule
Drug & Contraindicated condition \\
\midrule
Ketoconazole & torsades de pointes (QT prolongation) \\
Carbidopa & myocardial infarction \\
Diflunisal & hypertension \\
Desvenlafaxine & angle-closure glaucoma \\
Lisinopril & neonatal jaundice \\
\bottomrule
\end{tabular}
\caption{Contraindications that prose misses and the whole label names.}
\label{tab:labelonly}
\end{table}

These facts share one pattern. Each is a warning about a specific patient group or a rare harm: an arrhythmia, a heart attack, a rise in blood pressure, an eye emergency, a risk to a newborn. Prose about these drugs describes what they treat and how they work. The label lists who is at risk.

\subsection{Questions That Routing Fixes}

\Cref{tab:routefix} lists contraindication questions that all three 7-8B models answer correctly with the label page and none answers correctly with the prose page. In the sumatriptan question, for example, all three models pick \emph{coronary atherosclerosis}, a related heart disease, with the prose page. The label page names the answer, and all three models pick it.

\begin{table}[h]
\centering
\footnotesize
\setlength{\tabcolsep}{3pt}
\begin{tabular}{ll}
\toprule
Drug & Contraindicated condition \\
\midrule
Sumatriptan & Prinzmetal angina \\
Acetylcysteine & asthma \\
Apomorphine & orthostatic hypotension \\
Bimatoprost & uveitis \\
Fenoprofen & kidney disease \\
Levocetirizine & kidney disease \\
\bottomrule
\end{tabular}
\caption{Contraindication questions that all three 7-8B models answer correctly only with the label page.}
\label{tab:routefix}
\end{table}

\subsection{Errors and Edge Cases}

\paragraph{Combination labels.}
PrimeKG lists Crohn disease as a contraindication of tretinoin, and the label we retrieve names it. The label is for a gel that combines clindamycin with tretinoin, and the warning about inflammatory bowel disease comes from clindamycin. Our measurement counts this as named, as a retriever would. A clinical agent should check which ingredient a warning belongs to before it cites it.

\paragraph{Names in another sense.}
For histrelin, the label names \emph{hypertension} only inside \emph{idiopathic intracranial hypertension}, a different condition. For norepinephrine, a sentence says that alpha blockers treat hypertension. The matcher counts both as named. Three of four annotators mark the norepinephrine sentence as a mention with no link, and the histrelin sentence splits them evenly (\cref{app:relcheck}). The relation check exists to remove such cases.

\paragraph{Lists without verbs.}
Many label sentences are items in a list of adverse reactions, such as \emph{``Hematological: agranulocytosis, thrombocytopenia''}. The keyword check misses them because they contain no relation word. The judge and the annotators read them as warnings, because the section heading gives the relation. This is the main reason the keyword check agrees poorly with the annotators.

\paragraph{Approval fragments.}
Some indication sentences are fragments of a list of approved uses whose verb sits in an earlier sentence, such as doripenem and pyelonephritis. The judge sometimes reads these as mentions with no link. These errors remove real indications, so they make the measured gap smaller, not larger.